\documentclass{article}

\PassOptionsToPackage{hyphens}{url}
\usepackage[preprint]{neurips_2026}

\usepackage[utf8]{inputenc}
\usepackage[T1]{fontenc}
\usepackage{hyperref}
\usepackage{url}
\usepackage{booktabs}
\usepackage{amsfonts}
\usepackage{nicefrac}
\usepackage{microtype}
\usepackage{xcolor}
\usepackage{amsmath}
\usepackage{multirow} 
\usepackage{enumitem}
\usepackage{graphicx}
\usepackage{subcaption}
\usepackage{adjustbox}
\usepackage{wrapfig}
\usepackage{siunitx}
\usepackage{tabularx}
\usepackage{array}
\usepackage{xurl}
\usepackage{etoc}
\usepackage{placeins}

\hypersetup{colorlinks=true,allcolors=black}

\title{\textsc{CarCrashNet}: A Large-Scale Dataset and Hierarchical Neural Solver for Data-Driven Structural Crash Simulation}

\author{
  Mohamed Elrefaie\thanks{Corresponding author: \texttt{mohamed.elrefaie@mit.edu}} \\
  Department of Mechanical Engineering \\
  MIT \\
  Cambridge, MA, USA \\
  \And
  Dule Shu \\
  Future Product Innovation \\
  Toyota Research Institute \\
  Los Altos, CA, USA \\
  \AND
  Matt Klenk \\
  Future Product Innovation \\
  Toyota Research Institute \\
  Los Altos, CA, USA \\
  \And
  Faez Ahmed \\
  Department of Mechanical Engineering \\
  MIT \\
  Cambridge, MA, USA
}

\begin{document}

\maketitle

\begin{abstract}
Crash simulation is a cornerstone of modern vehicle development because it reduces the need for costly physical prototypes, accelerates safety-driven design iteration, and increasingly supports virtual testing workflows. At the same time, modeling structural crash mechanics remains exceptionally challenging: the response is governed by nonlinear contact, large deformation, material plasticity, failure, and complex multi-body interactions evolving over space and time on high-resolution finite-element meshes. While recent advances in scientific machine learning have shown that large-scale, high-fidelity datasets can drive major progress in physics-based prediction, structural crash simulation still lacks an open, validated, and machine-learning-ready benchmark of comparable scope. We introduce \textsc{CarCrashNet}, a public high-fidelity open-source benchmark for data-driven structural crash simulation. \textsc{CarCrashNet} combines component-scale and full-vehicle simulations in a multi-modal format, including more than 14{,}000 bumper-beam pole-impact simulations with varying geometry, materials, and boundary conditions, together with 825 full-vehicle crash simulations built from three industry-standard vehicle models of increasing structural complexity: Dodge Neon, Toyota Yaris, and Chevrolet Silverado. To establish the reliability of the benchmark, we validate our open-source finite-element workflow based on OpenRadioss against both experimental crash data and the commercial solver Ansys LS-DYNA. We also introduce \textsc{CrashSolver}, a machine-learning model designed for full-vehicle crash prediction from high-resolution finite-element crash data. We further perform extensive benchmarking across the released datasets and evaluate \textsc{CrashSolver} against state-of-the-art geometric deep learning and transformer-based neural solvers. Our results position \textsc{CarCrashNet} as a foundation for reproducible research in structural simulation, crashworthiness modeling, and AI-driven virtual crash testing. The dataset is available at \url{https://github.com/Mohamedelrefaie/CarCrashNet}.
\end{abstract}

\section{Introduction}
\label{sec:introduction}

Crash simulation is ultimately a safety problem, not only a computational one.
Small structural decisions, including millimeters of sheet thickness, local
load-path geometry, spot-weld placement, or rail stiffness, can change intrusion,
deceleration, and energy absorption during an impact. The societal stakes are
large: the National Highway Traffic Safety Administration (NHTSA), part of the
U.S. Department of Transportation, reported that motor-vehicle crashes cost
American society%
\footnote{NHTSA press release, ``Traffic Crashes Cost America Billions in
2019'': \url{https://www.nhtsa.gov/press-releases/traffic-crashes-cost-america-billions-2019}.}
\$340~billion in 2019, corresponding to crashes that killed an estimated
36{,}500 people, injured 4.5~million, and damaged 23~million vehicles; when
quality-of-life valuations are included, the total societal harm was nearly
\$1.4~trillion. The engineering cost is also substantial. Toyota reported in 2010 that it
conducted more than 1{,}600 physical vehicle crash tests per year across three
facilities, with each test costing about \$30{,}000 and requiring 11 working
days to plan and execute.%
\footnote{Toyota press release, ``Our Point of View: Anatomy of a Test Crash'':
\url{https://pressroom.toyota.com/our-point-of-view-anatomy-of-a-test-crash/}.}
More recently, Volvo reported that its Safety Centre crashes at least one
brand-new Volvo per day, performs tests beyond regulatory requirements, and uses
thousands of computer simulations before physical testing.%
\footnote{Volvo Cars press release, ``Two Decades in the Service of Saving Lives:
Volvo Cars Safety Centre Celebrates 20 Years'':
\url{https://www.volvocars.com/us/media/press-releases/E580236F0D12088A/}.}
These costs make high-fidelity simulation indispensable, but simulation data
must be validated, diverse, and machine-learning-ready before it can support
trustworthy surrogate modeling and virtual crash testing.

These challenges make crash simulation a particularly demanding but valuable
domain for scientific machine learning, where validated high-fidelity data could
enable fast surrogate models without discarding the mechanics that govern impact response. Recent developments in AI for physics and scientific machine learning are showing
increasingly promising results across a wide range of domains, including fluid
mechanics, weather and climate modeling, catalysis, and engineering design.
This progress is being driven by two complementary lines of work.  On the
modeling side, neural operators, operator-learning architectures, graph-based
simulators, transformer-style PDE solvers, and PDE foundation models have
expanded the range of physical systems that can be learned from data
\citep{kovachki2023neuraloperator,li2020fourier,lu2021deeponet,sanchez2020learning,pfaff2021meshgraphnets,wu2024transolver,poseidon2024,subramanian2023foundation,hassan2025bubbleformer, chen2025tripnet}.
In parallel, public datasets and benchmarks such as DrivAerNet++ \citep{elrefaie2024drivaernetpp},
DrivAerNet \citep{elrefaie2025drivaernet, elrefaie2024drivaernetConf}, PDEBench \citep{takamoto2022pdebench}, BlendedNet \citep{sung2025blendednet},
BlendedNet++ \citep{sung2025blendednetpp},
AirfRANS \citep{bonnet2022airfrans}, CFDBench \citep{luo2023cfdbench},
LagrangeBench \citep{toshev2023lagrangebench}, The Well \citep{well2024},
APEBench \citep{apebench2024}, PINNacle \citep{pinnacle2024},
WeatherBench \citep{rasp2020weatherbench},
CarBench \citep{elrefaie2025carbench}, 
ClimateBench \citep{watsonparris2022climatebench}, WxC-Bench \citep{shinde2024wxcbench},
RealPDEBench \citep{hu2026realpdebench}, and OC20 \citep{chanussot2021oc20}
have shown that dataset scale, fidelity, and task diversity play a critical
role in the rapid development of scientific ML.

Among these factors, the dataset itself is foundational. Large-scale,
high-fidelity, and well-curated data improves predictive performance and
determines whether models generalize across geometries, boundary conditions,
discretizations, and physical regimes. This is especially important in
engineering, where models must remain robust under changing designs and loading
conditions. The most useful scientific datasets are also increasingly
\emph{multi-modal}, combining geometry, state fields, scalar metrics, metadata,
and sometimes paired simulation--measurement observations
\citep{elrefaie2024drivaernetpp,shinde2024wxcbench,hu2026realpdebench}.

Despite this momentum, structural mechanics---and crash simulation in
particular---still lacks a public, open-source, high-fidelity benchmark of the
kind that has accelerated progress in neighboring scientific ML domains. Public
finite-element vehicle models do exist through sources such as CCSA and
NHTSA~\citep{ccsa_models,nhtsa_models},
and commercial solvers such as Ansys LS-DYNA remain the industrial standard for
nonlinear crash analysis~\citep{ansys_lsdyna}, but the field still lacks a large-scale,
machine-learning-ready, validated benchmark spanning both component-level and
full-vehicle crash simulation. This
gap is increasingly important as virtual crash testing becomes more central to
modern vehicle development and approval pipelines. For example, BMW recently
reported a hybrid homologation workflow in which selected physical crash tests
can be replaced by officially recognized virtual simulations while maintaining
the same safety standard\footnote{BMW Group, ''Hybrid homologation of the future'': 
\url{https://www.bmwgroup.com/en/news/general/2026/hybrid-homologation.html}.}. In such settings,
validation is not optional: dataset credibility depends not only on simulation
scale, but also on agreement with trusted commercial solvers and, ultimately,
physical experiments.

While the recent work of \citep{nabian2025automotive} is an important early
step toward machine-learning-accelerated crash dynamics, its scope is
fundamentally different from full-vehicle crash prediction. Their study focuses
on a simplified Body-in-White (BIW) assembly rather than a complete vehicle model, whereas
full-vehicle crash simulation introduces additional complexity from closures,
interior parts, suspension, wheels, powertrain-adjacent components, contact
interfaces, and heterogeneous material systems. Moreover, the dataset comprises only
150 LS-DYNA simulations, varies thickness distributions over a limited set of
components, and does not appear to include broader boundary-condition variation
or multiple vehicle architectures. Since the dataset is not publicly released,
the results are difficult to reproduce or extend, limiting its utility as a
community benchmark for data-driven crash simulation.

\begin{figure}[h!]
    \centering
    \includegraphics[width=\linewidth]{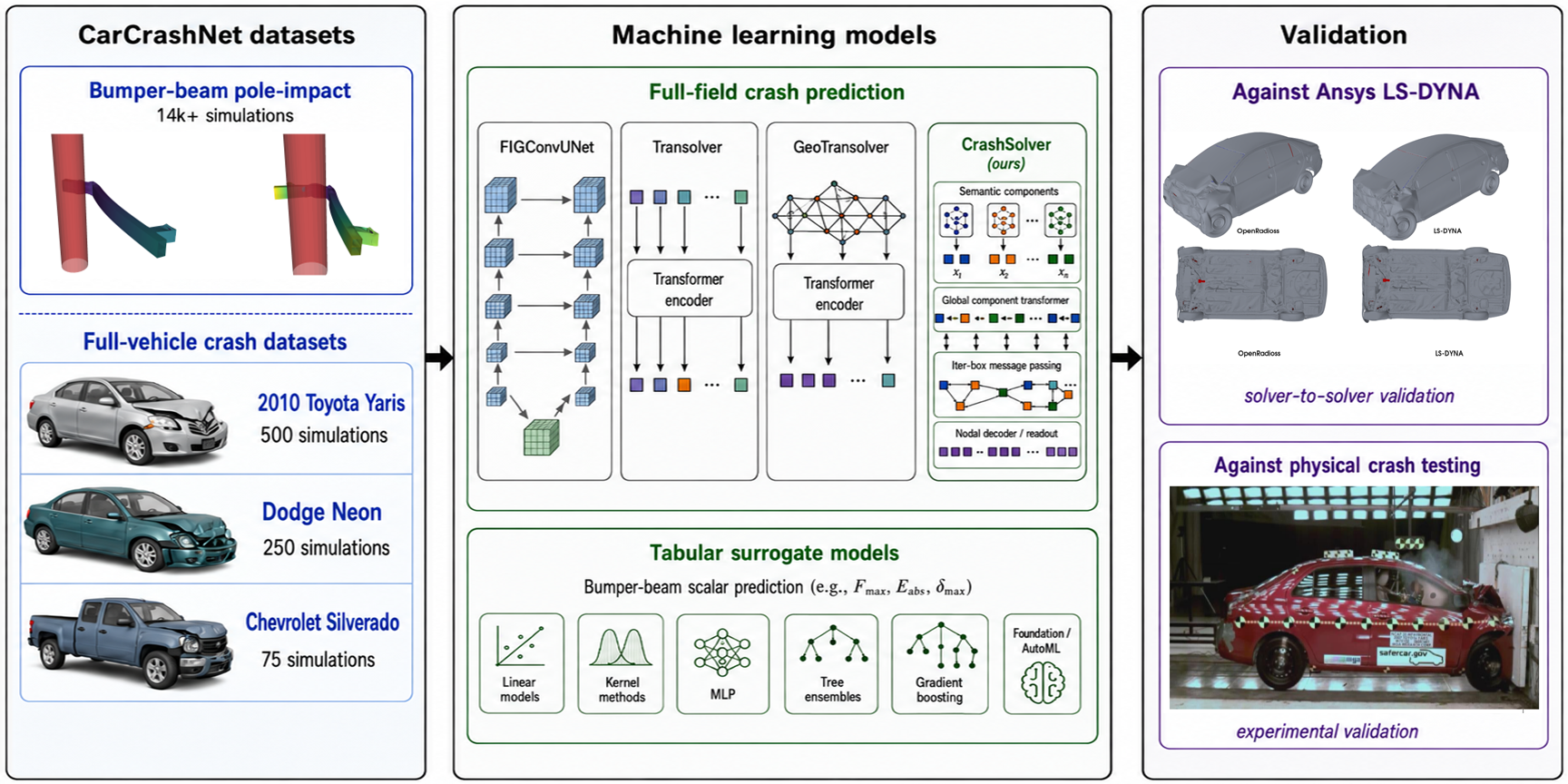}
    \caption{
    Overview of our \textsc{CarCrashNet} framework. Left: the released datasets include a large-scale bumper-beam pole-impact dataset with more than 14k simulations and three full-vehicle crash datasets based on the Toyota Yaris sedan, Dodge Neon, and Chevrolet Silverado. Middle: the machine learning tasks considered in this work include full-field crash prediction using FIGConvUNet, Transolver, GeoTransolver, and our proposed CrashSolver, as well as tabular surrogate modeling for crashworthiness quantities. Right: the dataset generation pipeline is validated through solver-to-solver comparison against Ansys LS-DYNA and against physical crash-testing references.}
    \label{fig:carcrashnet_overview}
\end{figure}

To address these gaps, we introduce \textsc{CarCrashNet} (see Fig.~\ref{fig:carcrashnet_overview}), which to the best of our
knowledge is the first public high-fidelity open-source dataset for
crash-oriented structural simulation that combines validated component-scale and
full-vehicle simulations in a machine-learning-ready format. Our contributions
are three-fold:
\begin{enumerate}
    \item We validate the open-source finite-element analysis workflow
    OpenRadioss against physical crash-testing 
    and the commercial industry-standard solver Ansys LS-DYNA, establishing a
    reproducible and physically grounded foundation for open structural crash
    simulation research.

    \item We release the first large-scale public dataset for structural crash
    simulation and crash testing at two levels of fidelity and complexity:
    (i) a component-scale bumper-beam dataset (14k samples) for frontal pole impact with
    varying geometry, material, and boundary conditions, and
    (ii) a full-vehicle crash dataset (825 samples) built from three industry-standard
    vehicle models with increasing structural and geometric diversity, namely Toyota Yaris, Dodge Neon, and Chevrolet Silverado.

    \item We introduce \textsc{CrashSolver}, a hierarchical machine-learning model for full-vehicle
    crash prediction and perform an extensive benchmark across the released
    datasets, comparing a broad suite of state-of-the-art models and showing that the proposed method is a strong structured baseline for this challenging
    four-dimensional structural simulation task.
\end{enumerate}

Overall, \textsc{CarCrashNet} aims to provide the missing open dataset needed
to advance data-driven structural mechanics, in the same way that large public
datasets have accelerated progress in fluid dynamics, weather, and climate
forecasting, and atomistic simulation. By combining solver validation,
experimental grounding, dataset diversity, and multi-modal representations, the
dataset is designed to support future work on surrogate modeling, inverse
design, foundation models for structural simulation, and trustworthy virtual
testing pipelines.

\paragraph{Paper organization.}
Section~\ref{sec:datasets} introduces \textsc{CarCrashNet} and its
bumper-beam and full-vehicle simulation campaigns. Section~\ref{sec:crashsolver_benchmarking}
presents \textsc{CrashSolver} and the main benchmark results, and
Sections~\ref{sec:conclusion} and~\ref{sec:limitations_future_work} summarize
findings, limitations, and responsible use. The appendix provides related work
(Appendix~\ref{sec:related_work}), solver validation
(Appendix~\ref{sec:solver_validation}), dataset-generation details
(Appendices~\ref{app:vehicle_baseline_models} and~\ref{sec:bumper-dataset}),
CrashSolver ablations and concurrent evaluation details
(Appendices~\ref{app:crashsolver_ablation_studies} and~\ref{app:concurrent_suv_eval}),
statistical-significance analysis (Appendix~\ref{app:statistical_significance}),
and bumper-beam ML benchmarks (Appendix~\ref{sec:ml-benchmark}).

\section{Crash Simulation Datasets}
\label{sec:datasets}

In this section, we introduce the two datasets that form \textsc{CarCrashNet}: a component-scale bumper-beam pole-impact dataset with more than 14{,}000 validated simulations, and a vehicle-scale frontal crash dataset spanning Dodge Neon, Toyota Yaris, and Chevrolet Silverado. Together, they provide complementary benchmarks for controlled design-space exploration and realistic full-vehicle crash prediction.

\subsection{Bumper-Beam Pole-Impact Crashworthiness Dataset}
\label{sec:bumper_beam_main}
The component-scale dataset is built around a frontal crash structure
consisting of a DP1000 bumper beam and DP600 crash boxes, where DP denotes
dual-phase steel, impacting a rigid cylindrical pole. This setting is intentionally chosen as an intermediate-scale
benchmark: it is simpler than a complete vehicle, but still contains the key
physics that make crash simulation challenging, including nonlinear contact,
large deformation, plasticity, local crushing, and energy absorption. The
dataset contains 14{,}742 finite-element OpenRadioss simulations sampled over
seven engineering design variables: impact velocity, crash-box thickness,
bumper-beam thickness, DP600 yield strength, DP1000 yield strength, pole
diameter, and lateral pole offset. More details on the finite-element model,
materials, boundary conditions, design-of-experiments setup, validation filters,
and released data products are provided in Appendix~\ref{sec:bumper-dataset}.

Each simulation provides scalar crashworthiness quantities, including peak pole
contact force, peak deceleration, peak internal energy, peak plastic work, and
kinetic-energy absorption. In addition, the released simulation outputs include
field-level trajectory data for geometry-aware and spatiotemporal learning. We
also use this dataset for a compact machine-learning benchmark on tabular
surrogate models, comparing linear models, kernel methods, neural networks,
tree ensembles, gradient-boosted trees, AutoML, and a tabular foundation model.
A detailed analysis of the benchmark setup, train/validation/test splits,
model suite, evaluation protocol, and results is provided in
Appendix~\ref{sec:ml-benchmark}.

\subsection{Vehicle-Scale Crash Simulation Dataset}
\label{sec:vehicle_campaigns}

\textsc{CarCrashNet} extends beyond component-level crash benchmarks by
including vehicle-scale explicit finite-element simulations for three public full-vehicle models: a Dodge Neon passenger car, a Toyota Yaris
passenger car, and a detailed Chevrolet Silverado pickup truck.

Relative to the independently run Ansys LS-DYNA reference, the detailed
OpenRadioss baseline differs by 7.2\% in CFC60 peak wall force, 2.6\% in
wall-force duration, and 0.5\% in peak internal energy. Relative to the
published physical-test scale, it matches the impact speed within 0.2\%,
overpredicts peak wall force by approximately 14.6\%, and underpredicts the
wall-force duration by approximately 19.6\%; these comparisons support global
response agreement while identifying contact and pulse-shape quantities as
solver-sensitive. Extensive solver validation against experimental references and Ansys LS-DYNA is provided in Section~\ref{sec:solver_validation}. 

The objective is to generate
machine-learning datasets that preserve realistic vehicle topology, material
heterogeneity, contact interactions, and transient deformation while remaining
structured enough for controlled surrogate-model evaluation.  We therefore
construct compact design-of-experiments (DoE) campaigns around validated
baseline models, instead of perturbing arbitrary mesh, material, and solver
parameters.

At the dataset level, each simulated case is represented as
\begin{equation}
  \mathcal{D}_c =
  \left(
    \boldsymbol{\xi}_c,\,
    \mathcal{G}_c,\,
    \mathcal{H}_c,\,
    \mathbf{y}_c,\,
    \mathbf{m}_c
  \right),
\end{equation}
where $\boldsymbol{\xi}_c$ is the design vector, $\mathcal{G}_c$ is the
time-resolved field trajectory exported to VTKHDF, $\mathcal{H}_c$ is the set
of scalar time histories extracted from the solver outputs, $\mathbf{y}_c$ is
the vector of reduced quantities of interest, and $\mathbf{m}_c$ contains case
metadata such as split labels, anchor flags, and source-campaign provenance.

\subsubsection{Simulation Formulation}
\label{sec:vehicle_campaigns:formulation}

\paragraph{Semi-discrete crash dynamics.}
All three campaigns are solved with the explicit transient-dynamics workflow in
OpenRadioss~\citep{openradioss_github,altair_radioss}. At the semi-discrete finite-element level, the vehicle response is
governed by
\begin{equation}
  \mathbf{M}\,\ddot{\mathbf{u}}(t) +
  \mathbf{f}_{\mathrm{int}}\!\left(\mathbf{u}(t), \dot{\mathbf{u}}(t);
  \boldsymbol{\theta}_{\mathrm{mat}}\right) +
  \mathbf{f}_{\mathrm{cont}}\!\left(\mathbf{u}(t), \dot{\mathbf{u}}(t)\right)
  =
  \mathbf{f}_{\mathrm{ext}}\!\left(t; \boldsymbol{\xi}\right),
  \label{eq:vehicle_explicit_dynamics}
\end{equation}
where $\mathbf{u}(t)$ is the nodal displacement field,
$\mathbf{M}$ is the assembled mass matrix,
$\mathbf{f}_{\mathrm{int}}$ is the internal force induced by the constitutive
response,
$\mathbf{f}_{\mathrm{cont}}$ is the nonlinear contact contribution, and
$\mathbf{f}_{\mathrm{ext}}$ is the load state induced by the crash setup.
Within a given campaign, the material parameter vector
$\boldsymbol{\theta}_{\mathrm{mat}}$ is inherited from the baseline model and
held fixed; only the low-dimensional design vector $\boldsymbol{\xi}$ is
varied.

\subsubsection{Design Variables and Sampling}
\label{sec:vehicle_campaigns:design}

All vehicle campaigns use the same abstract design vector,
\begin{equation}
  \boldsymbol{\xi}
  =
  \left[
    v,\,
    s_{\mathrm{front}},\,
    s_{\mathrm{rail}}
  \right],
  \label{eq:vehicle_design_vector}
\end{equation}
where $v$ is impact velocity, $s_{\mathrm{front}}$ scales front-support shell
thicknesses, and $s_{\mathrm{rail}}$ scales the lower-rail or subframe load
path. The velocity range is
$v\in[50,64]\,\si{\kilo\metre\per\hour}$ for all three campaigns.

\begin{table}[h!]
\centering
\footnotesize
\caption{Structural thickness design space used for the vehicle-scale DoE campaigns.}
\label{tab:vehicle_thickness_design_space}
\setlength{\tabcolsep}{3.5pt}
\renewcommand{\arraystretch}{1.08}
\begin{tabular*}{\textwidth}{@{\extracolsep{\fill}} l c c c c c @{}}
\toprule
\textbf{Vehicle} &
\shortstack{\textbf{Front}\\\textbf{support}} &
\shortstack{\textbf{Lower rail /}\\\textbf{subframe}} &
\shortstack{\textbf{Total edited}\\\textbf{groups}} &
\shortstack{\textbf{Thickness}\\\textbf{range}} &
\shortstack{\textbf{Percent}\\\textbf{range}} \\
\midrule
Dodge Neon          & 12 shell thickness groups & 15 shell thickness groups & 27 & $[0.9,1.1]\,t_0$ & $\pm 10\%$ \\
Toyota Yaris        & 27 shell thickness groups & 10 shell thickness groups & 37 & $[0.9,1.1]\,t_0$ & $\pm 10\%$ \\
Chevrolet Silverado & 29 shell thickness groups & 49 shell thickness groups & 78 & $[0.9,1.1]\,t_0$ & $\pm 10\%$ \\
\bottomrule
\end{tabular*}
\end{table}

For structural groups, the edited thickness for group $g$ and case $c$ is
\begin{equation}
  t_{g}^{(c)} = s_{g}^{(c)} t_{g}^{(0)}, \qquad
  s_g^{(c)} \in [0.9,1.1],
  \label{eq:vehicle_thickness_scaling}
\end{equation}
so selected front-end thicknesses vary within $\pm10\%$ of nominal. The number
of edited front-support and lower-rail/subframe groups for each vehicle is
summarized in Table~\ref{tab:vehicle_thickness_design_space}.

\begin{figure}[h!]
    \centering
    \includegraphics[width=\linewidth]{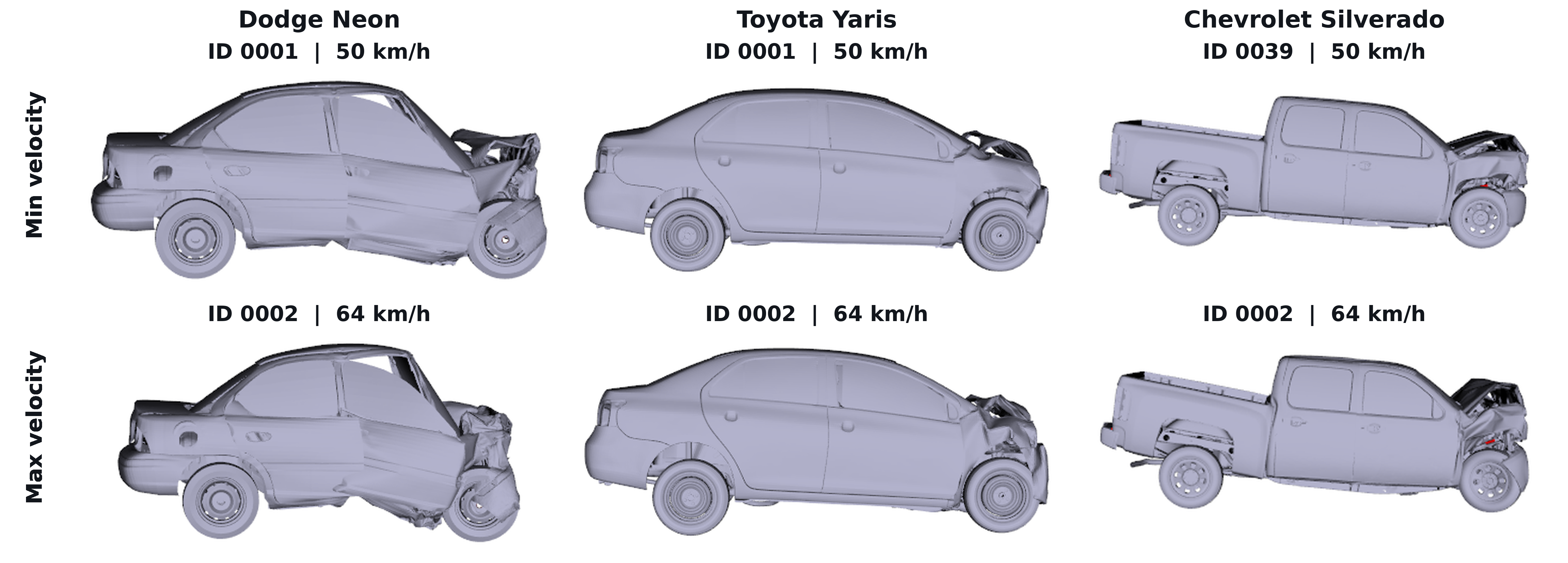}
    \caption{
      Representative low- and high-velocity crash cases for the three
      vehicle-scale models. The top row shows low-speed impact cases and
      the bottom row shows high-speed impact cases. Increasing velocity
      produces more severe front-end deformation while preserving clear
      vehicle-dependent differences in crush mode.
    }
    \label{fig:vehicle_velocity_comparison}
\end{figure}

The design variables were chosen to couple loading severity with physically
meaningful structural stiffness changes. Impact speed controls the initial
kinetic-energy scale and therefore the deformation regime, as illustrated by the
low- and high-speed examples in Fig.~\ref{fig:vehicle_velocity_comparison}.

\begin{figure}[h!]
    \centering
    \includegraphics[width=\linewidth]{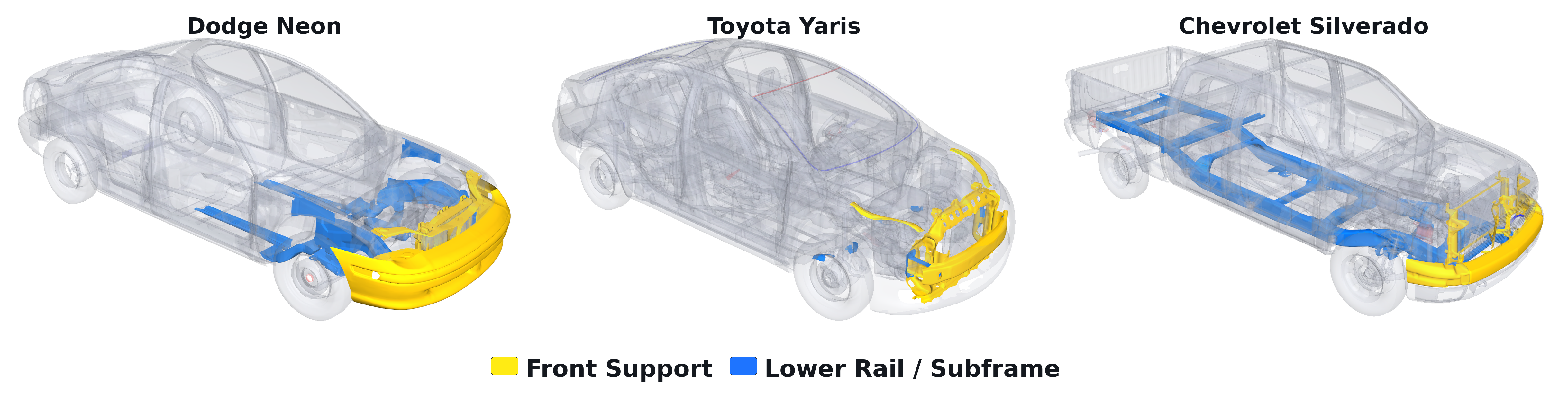}
    \caption{
      Structural regions edited by the vehicle-scale DoE. Highlighted front
      support and lower-rail/subframe groups define the thickness-scaling
      variables in Eq.~\eqref{eq:vehicle_design_vector}. The same abstract
      design variables are used across vehicles, but the underlying physical
      parts differ because each vehicle has a distinct frontal load path.
    }
    \label{fig:vehicle_design_parameters}
\end{figure}

The front-support group contains bumper, radiator-support, and front-end bracket
structures that control early contact and crush initiation, while the
lower-rail/subframe group contains longitudinal rails, frame supports,
suspension-frame attachments, and crossmembers that govern load transfer into
the main body or frame. These edited structural regions are highlighted in
Fig.~\ref{fig:vehicle_design_parameters}. This parameterization gives each
vehicle a low-dimensional design space while still targeting the dominant
frontal crash load path.

For the Yaris and Silverado campaigns, interior samples are generated by
Latin-hypercube sampling~\citep{mckay1979lhs} after reserving deterministic anchor cases at the
baseline, one-factor extrema, and design-space corners. For Neon, the first
75 cases establish the same anchor-backed pilot structure, while subsequent
batches use greedy maximin continuation~\citep{johnson1990minimax}. If $\mathcal{X}_k$ is the accumulated
normalized design set and $\mathcal{C}$ is a candidate pool, the next
continuation point is
\begin{equation}
  \mathbf{x}_{k+1}
  =
  \arg\max_{\mathbf{x}\in\mathcal{C}}
  \min_{\mathbf{z}\in\mathcal{X}_k}
  \left\lVert \mathbf{x}-\mathbf{z}\right\rVert_2 .
  \label{eq:vehicle_maximin}
\end{equation}
This expands training coverage without changing the fixed anchor and held-out
test structure. More details related to the baseline models and the design space can be found in Section~\ref{app:vehicle_baseline_models}.

\subsubsection{Released Fields, Histories, and Quantities of Interest}
\label{sec:vehicle_campaigns:fields}

\paragraph{Field trajectories.}
Each completed case is converted from raw solver files into a VTKHDF
partitioned dataset that preserves per-part topology and time-resolved state.
For case $c$, part block $p$, and timestep $t_n$, we represent the exported
field state as
\begin{equation}
\begin{aligned}
  \mathcal{G}_{c,p}(t_n) =
  \{&
    \mathbf{X}^{(0)},\,
    \mathbf{X}(t_n),\,
    \mathbf{U}(t_n),\,
    \dot{\mathbf{U}}(t_n),\,
    \sigma_{\mathrm{vm}}(t_n), \\
    &
    \bar{\varepsilon}_{\mathrm{p}}(t_n),\,
    e_{\mathrm{spec}}(t_n),\,
    \eta_{\mathrm{erode}},\,
    \mathrm{id}_{\mathrm{node}},\,
    \mathrm{id}_{\mathrm{elem}},\,
    \mathrm{id}_{\mathrm{part}}
  \}.
\end{aligned}
  \label{eq:vehicle_field_state}
\end{equation}
Concretely, the exported VTKHDF blocks store the mesh geometry as timestep-indexed
\texttt{Points}: the first frame provides the undeformed reference coordinates
$\mathbf{X}^{(0)}$, and later frames provide the deformed coordinates
$\mathbf{X}(t_n)$. The same blocks also store nodal point data including
displacement $\mathbf{U}(t_n)$, velocity $\dot{\mathbf{U}}(t_n)$, and node
identifier. Element-level data include von Mises equivalent stress, equivalent
plastic strain, specific internal energy, erosion or failure status, element
identifier, and part identifier. This representation therefore preserves both
the undeformed mesh and the time-resolved deformed meshes, while also exposing
the displacement fields used by downstream visualization, geometry-aware
learning, and spatiotemporal surrogate modeling.

\paragraph{Global and local time histories.}
In addition to field trajectories, each case retains the solver history
outputs. We denote the shared history interface
as
\begin{equation}
  \mathcal{H}_c(t) =
  \left\{
    \mathbf{F}_{\mathrm{wall}}(t),\,
    E_{\mathrm{kin}}(t),\,
    E_{\mathrm{int}}(t),\,
    E_{\mathrm{cont}}(t),\,
    E_{\mathrm{hg}}(t),\,
    a_j(t)
  \right\},
  \label{eq:vehicle_history_state}
\end{equation}
where $\mathbf{F}_{\mathrm{wall}}(t)$ is the rigid-wall reaction,
$E_{\mathrm{kin}}(t)$, $E_{\mathrm{int}}(t)$, $E_{\mathrm{cont}}(t)$, and
$E_{\mathrm{hg}}(t)$ are the kinetic, internal, contact, and hourglass
energies, and $a_j(t)$ denotes deck-specific local acceleration channels when
such history channels are present in the baseline model. The exact local
channels depend on the source deck, but the core global force and energy traces
are common across campaigns.

\paragraph{Reduced quantities of interest.}
From the time histories we compute scalar quantities of interest that are more
convenient for tabular learning, benchmarking, and validation:
\begin{align}
  F_{\mathrm{wall}}^{\max}
    &= \max_t \left\lVert \mathbf{F}_{\mathrm{wall}}(t) \right\rVert_2, \\
  E_{\mathrm{int}}^{\max}
    &= \max_t E_{\mathrm{int}}(t), \\
  \eta_{\mathrm{KE}}
    &= 1 - \frac{E_{\mathrm{kin}}(T_f)}{E_{\mathrm{kin}}(0)}, \\
  a_j^{\max}
    &= \max_t \left| a_j(t) \right|, \\
  t_1
    &= \inf\left\{t:\left\lVert \mathbf{F}_{\mathrm{wall}}(t) \right\rVert_2
    > 0.03\,F_{\mathrm{wall}}^{\max}\right\}, \\
  t_2
    &= \sup\left\{t:\left\lVert \mathbf{F}_{\mathrm{wall}}(t) \right\rVert_2
    > 0.03\,F_{\mathrm{wall}}^{\max}\right\}, \\
  T_{\mathrm{imp}}
    &= t_2 - t_1.
  \label{eq:vehicle_qoi}
\end{align}
Here $T_f$ denotes the final simulated time.
These reduced quantities are not intended to replace the field trajectories;
they provide a compact audit and benchmarking layer on top of the full
geometry- and state-resolved outputs.

For spatiotemporal surrogate modeling, the released vehicle campaigns support a
Lagrangian field-prediction interface: node identity and element connectivity
are fixed across output frames within each case, while deformation is represented
by nodal motion. In the most complete form, a field predictor receives the
reference nodal coordinates, retained mesh or surface connectivity, FE part
identifiers or mapped component labels, and the design vector, then predicts the
future displacement trajectory,
\begin{equation}
  f_{\theta}
  \left(
    \mathbf{X}^{(0)},
    \mathcal{E},
    \mathbf{p},
    \boldsymbol{\xi}
  \right)
  =
  \hat{\mathbf{U}}^{(1:n)}
  \in \mathbb{R}^{n \times N \times 3},
  \label{eq:vehicle_learning_target}
\end{equation}
where $\mathbf{X}^{(0)}$ is the undeformed nodal coordinate matrix,
$\mathcal{E}$ is the retained mesh or surface connectivity used by the learning
model, $\mathbf{p}$ contains FE part identifiers or their mapped
semantic-component labels depending on the model variant, and
$\hat{\mathbf{U}}^{(1:n)}$ is the predicted displacement sequence. Deformed
positions are recovered as
$\hat{\mathbf{X}}^{(t)} = \mathbf{X}^{(0)} + \hat{\mathbf{U}}^{(t)}$.

\section{CrashSolver: Full-Vehicle Crash Field Prediction}
\label{sec:crashsolver_benchmarking}

In this work we introduce \textsc{CrashSolver}, a learning architecture for
full-vehicle crash finite-element field prediction.  
CrashSolver is a hierarchical ML surrogate architecture designed around complete full-vehicle crash simulations.
Given the undeformed
mesh at $t=0$, CrashSolver
predicts the future nodal displacement trajectory over the crash event; nodal
positions are recovered as $\mathbf{X}^{(0)}+\hat{\mathbf{U}}^{(t)}$.  The task
is difficult because the mesh contains hundreds of thousands of nodes,
deformation is highly localized near the impact load path, and long-range
structural interactions propagate through rails, bumper systems, engine-bay
supports, cabin members, and subframe components.

\begin{figure}[h!]
    \centering
    \includegraphics[width=\linewidth]{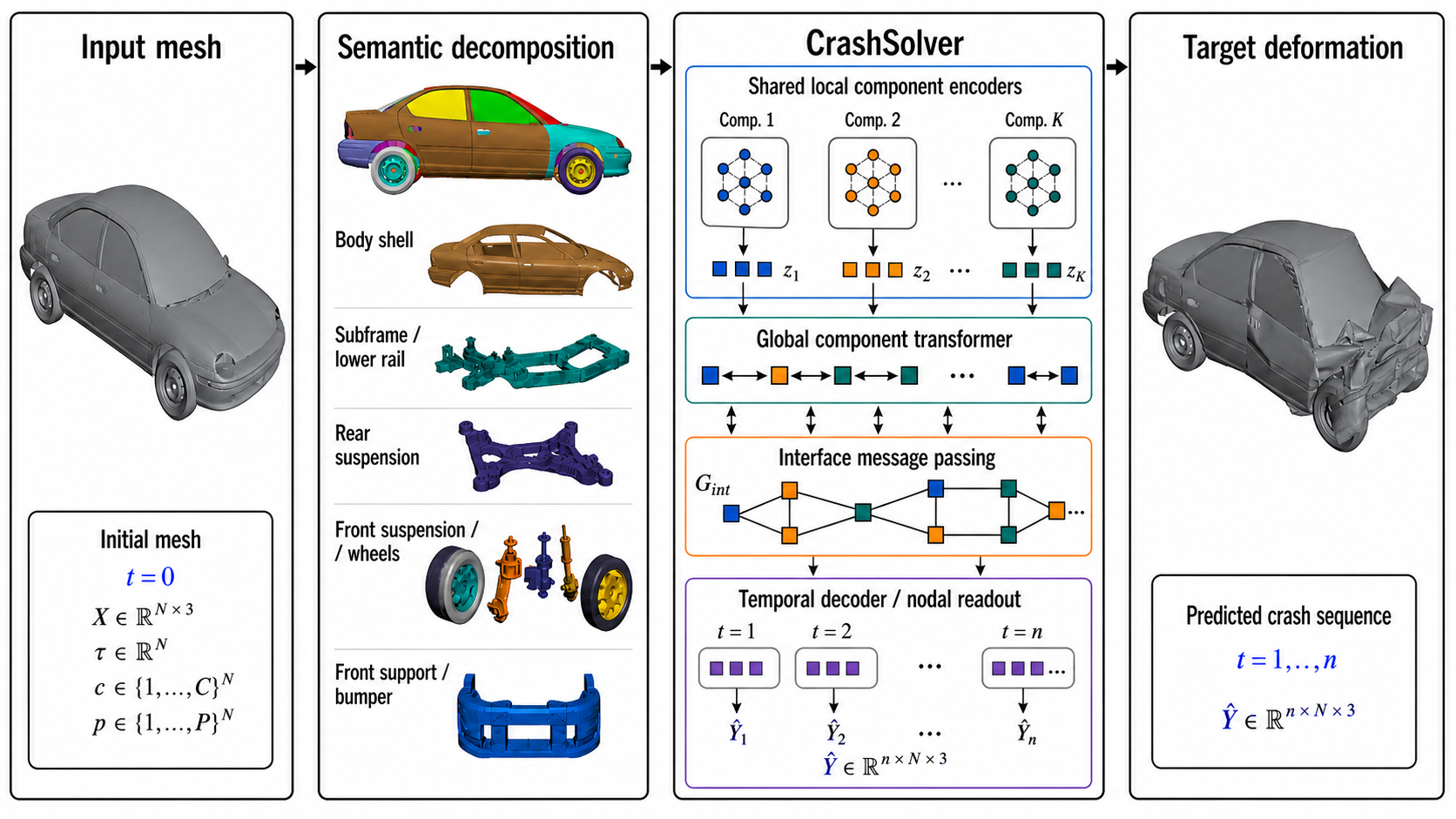}
\caption{
Overview of CrashSolver for full-vehicle crash prediction. Starting from the undeformed input mesh, the vehicle is decomposed into semantic structural components, which are processed by shared local component encoders, a global component transformer, and interface message passing, followed by a temporal decoder that predicts the future crash sequence. Here, $N$ denotes the number of mesh nodes, $t$ is the time index, and $n$ is the number of predicted future time steps. The input features are the initial nodal coordinates $X \in \mathbb{R}^{N \times 3}$, nodal thickness or gauge feature $\tau \in \mathbb{R}^{N}$, semantic component labels $c \in \{1,\dots,C\}^{N}$, and dense finite-element part labels $p \in \{1,\dots,P\}^{N}$, where $C$ is the number of semantic components and $P$ is the number of distinct FE parts. The latent vector $z_k$ denotes the learned embedding of component $k$, and $G_{\mathrm{int}}$ denotes the cross-component interface graph used for message passing. The output is the predicted nodal displacement sequence $\hat{\mathbf{Y}} = \hat{\mathbf{U}} \in \mathbb{R}^{n \times N \times 3}$, where $\hat{\mathbf{U}}^{(t)}$ is the predicted nodal displacement at future time step $t$.}    \label{fig:crashsolver}
\end{figure}

\paragraph{CrashSolver design.}
CrashSolver uses the finite-element part hierarchy as an inductive bias. The
reader maps FE \texttt{PART\_ID} values into semantic structural groups such
as bumper, rails, radiator support, shock housings, subframe, engine bay,
cabin floor, rocker, pillars, and exterior panels. Each component is encoded
with a shared lightweight geometry-aware attention block, so local deformation
is learned on smaller structural token sets instead of through a single
monolithic attention layer over the full vehicle. Component summaries are
then mixed by a global transformer attention block~\citep{vaswani2017attention},
and mesh-derived interface message passing exchanges latent features across
component boundaries. This design preserves the high-resolution nodal
deformation field while giving the network an explicit representation of crash
load paths, as illustrated in Fig.~\ref{fig:crashsolver}.

\paragraph{Baselines and protocol.}
We compare our model against state-of-the-art transformer-based neural solvers as well as geometric deep learning models.  Transolver
\citep{wu2024transolver} and GeoTransolver~\citep{adams2025geotransolver}
operate on the retained vehicle surface as one point set and use learned
physics/geometric slicing attention; GeoTransolver additionally uses the
geometry-aware local feature path exposed by the PhysicsNeMo implementation~\citep{physicsnemo_github}. We also
include a FIGConvUNet baseline~\citep{choy2025factorized} as a point-cloud
convolutional U-shaped model on the retained surface nodes.  For
completed leaderboard rows, all models use the same train/validation/test
split and the same retained node subset.

The main-paper evaluation reports three \textsc{CarCrashNet} vehicle-scale
benchmarks.  The Toyota Yaris benchmark has a fixed vehicle topology and an 80/10/10\% train/validation/test partition after quality-control (QC) filtering of the
500-case design-of-experiments campaign, with 50 held-out test runs; this
isolates architecture effects on one full-vehicle mesh.  The Dodge Neon
benchmark provides a second passenger-car topology and uses an
80/10/10\% partition with 25 held-out test runs, testing whether the
same model family remains effective when the mesh, part hierarchy, materials,
and load-path layout change.  The Silverado benchmark provides a larger
pickup-truck scaling test over the 75-case campaign, with a 56/4/15
train/validation/test split (approximately 75/5/20\%). We report mean absolute position error, RMSE,
relative $L_2$ position error, displacement-relative error when available, and
time-local RMSE summaries.  Lower is better for all metrics.  

\begin{table}[h!]
  \centering
  \caption{
    Full-vehicle crash prediction performance on the unseen hidden test set for
    each CarCrashNet vehicle dataset. Lower is better. Rows within each
    dataset are ranked by mean RMSE. Here $L_2^x$ denotes relative position
    error and $L_2^u$ denotes relative displacement error. Best values are highlighted in bold.}
    
  \label{tab:full_vehicle_hidden_test}
  \scriptsize
  \setlength{\tabcolsep}{2.5pt}
  \renewcommand{\arraystretch}{1.08}
  \resizebox{\linewidth}{!}{%
  \begin{tabular}{c l c c c c c}
    \toprule
    Rank &
    Model &
    RMSE (mm) &
    MAE (mm) &
    Relative $L_2^x$ &
    Relative $L_2^u$ &
    RMSE@60ms / RMSE$_T$ (mm) \\
    \midrule

        \multicolumn{7}{c}{Dodge Neon dataset (25 hidden test simulations)} \\
    \midrule
    1 & \textbf{CrashSolver}    & \textbf{32.763} & \textbf{18.036} & \textbf{0.02499} & \textbf{0.08837} & \textbf{50.904} \\
    2 & Transolver    & 33.947 & 18.678 & 0.02589 & 0.09148 & 52.759 \\
    3 & FIGConvUNet   & 34.044 & 18.850 & 0.02597 & 0.09196 & 53.330 \\
    4 & GeoTransolver & 34.403 & 18.973 & 0.02628 & 0.09349 & 52.660 \\

    \midrule

    \multicolumn{7}{c}{Toyota Yaris dataset (50 hidden test simulations)} \\

    \midrule
    1 & \textbf{CrashSolver}   & \textbf{21.769} & 13.507 & \textbf{0.01537} & \textbf{0.09043} & 22.714 \\
    2 & GeoTransolver & 21.773 & \textbf{13.359} & \textbf{0.01537} & 0.09059 & \textbf{22.677} \\
    3 & FIGConvUNet   & 21.910 & 13.576 & 0.01547 & 0.09105 & 22.821 \\
    4 & Transolver    & 22.583 & 14.049 & 0.01594 & 0.09391 & 23.642 \\

    \midrule
    \multicolumn{7}{c}{Chevrolet Silverado dataset (15 hidden test simulations)} \\
    \midrule
    1 & \textbf{CrashSolver}    & \textbf{61.536} & \textbf{37.753} & \textbf{0.03143} & \textbf{0.17069} & \textbf{61.078} \\
    2 & GeoTransolver & 79.230  & 45.366 & 0.04049 & 0.21844 & 79.410 \\
    3 & Transolver    & 83.971  & 47.510 & 0.04291 & 0.23184 & 84.008 \\
    4 & FIGConvUNet   & 102.747 & 62.405 & 0.05248 & 0.28432 & 84.863 \\

    \bottomrule
  \end{tabular}%
  }
\end{table}

Table~\ref{tab:full_vehicle_hidden_test} summarizes full-vehicle crash prediction
performance on the unseen hidden test sets of the three \textsc{CarCrashNet}
vehicle datasets. Across all vehicles, CrashSolver achieves the lowest mean
RMSE, indicating that the semantic component hierarchy, interface message
passing, and part-aware conditioning provide a consistent advantage for
full-vehicle deformation prediction. We provide uncertainty and significance analysis in
Appendix~\ref{app:statistical_significance}. On the Toyota Yaris dataset, CrashSolver
and GeoTransolver are nearly tied, with CrashSolver slightly improving RMSE and
relative displacement error, while GeoTransolver gives the lowest MAE and
RMSE at 60\,ms. This suggests that the Yaris split is relatively well captured
by both geometric transformer-style models and the proposed hierarchical
architecture.

The advantage of CrashSolver becomes clearer as the benchmark shifts to more
challenging vehicle settings. On the Dodge Neon dataset, which changes the
vehicle topology and frontal load path while retaining the same abstract design
variables, CrashSolver ranks first across all reported metrics, including RMSE,
MAE, relative position error, relative displacement error, and final-frame
RMSE. The largest separation appears on the Chevrolet Silverado dataset, where
the model must handle a larger pickup-truck design with different load-path structure and higher geometric complexity. In this setting,
CrashSolver reduces mean RMSE from 79.230\,mm for the best competing baseline
to 61.536\,mm, showing that the proposed component-aware representation is most
beneficial when the structural system becomes larger and more heterogeneous.
Overall, these results indicate that CrashSolver provides the strongest and most
consistent full-vehicle crash surrogate across the hidden test sets.  Extensive ablation studies are
reported in Section~\ref{app:crashsolver_ablation_studies}, and additional
evaluations are provided in Section~\ref{app:concurrent_suv_eval}.

\section{Conclusion}
\label{sec:conclusion}

We introduced \textsc{CarCrashNet}, a 6.65\,TB high-fidelity open dataset for
data-driven structural crash simulation. The dataset combines a large
component-scale bumper-beam pole-impact corpus with vehicle-scale full-vehicle
crash simulations across three finite-element baselines: the Dodge Neon, Toyota
Yaris, and Chevrolet Silverado. By varying impact velocity and structurally
meaningful front-end thickness parameters, the dataset captures both controlled
component-level behavior and more complex full-vehicle deformation modes. We
also establish a validation pathway for open crash simulation by comparing
OpenRadioss against the industry-standard Ansys LS-DYNA reference and available
experimental crash data, showing that the open-source workflow is suitable for
global-response dataset generation.

Beyond dataset generation, \textsc{CarCrashNet} provides machine-learning-ready
field trajectories, scalar histories, reduced crashworthiness metrics, and
metadata for reproducible evaluation. We further introduced \textsc{CrashSolver},
a hierarchical neural solver that uses semantic structural components,
part-aware conditioning, global interaction modeling, and interface message
passing to predict full-vehicle crash deformation fields. Across the Dodge Neon,
Toyota Yaris, and Chevrolet Silverado hidden test sets, \textsc{CrashSolver}
achieves the strongest overall performance, with the largest gains on the more
complex Silverado pickup-truck benchmark. Together, these contributions provide
a foundation for full-field crash surrogate modeling, design optimization,
cross-vehicle generalization, and trustworthy AI-assisted virtual crash testing.

\bibliographystyle{plainnat}
\bibliography{references}

\clearpage

\appendix

\part*{Appendix}
\addcontentsline{toc}{part}{Appendix}

\etocsettocstyle{\section*{Appendix contents}}{}
\etocsetnexttocdepth{subsection}
\localtableofcontents

\section{Related Work}
\label{sec:related_work}

Machine learning for crash-oriented finite element analysis (FEA) has evolved from classical surrogate modeling toward mesh-aware prediction and solver-integrated hybridization. In automotive crashworthiness, this transition is driven by the high cost of explicit dynamic simulations, which remain the standard tool for analyzing bumper beams, crash boxes, thin-walled absorbers, and vehicle body structures under impact loading \citep{marzbanrad2009design,mohd2025crashworthiness,salanke2025review,mukhudwana2025automotive}. These engineering studies also explain why crash learning is difficult: the response is highly nonlinear, transient, path-dependent, and sensitive to geometry, thickness, material choice, and boundary conditions.

A first family of methods treats the simulator as a black box and learns \emph{offline surrogates} from simulation data. A representative example is \citep{kohar2021machine}, who propose a voxelization procedure for unstructured FE data, use a 3D-CNN autoencoder to obtain latent representations, and then apply an LSTM to predict force--displacement response and mesh deformation. This line of work is important because it moves beyond scalar metamodels while still preserving a standard train-once, infer-many surrogate workflow. Relatedly, \citep{borse2023machine} combine FE simulations, reinforcement learning, and GAN-generated synthetic data for crash-box design optimization, illustrating how learned surrogates can be embedded directly into design search. A more task-specific example is \citep{liu2025intelligent}, who predict local collision damage from FE-generated simulation datasets rather than reconstructing full deformation fields. Offline methods are attractive because they are easy to integrate into optimization loops and can be effective even when the design variables are low-dimensional, but they generally remain limited in geometric generalization and solver compatibility.

A second family comprises \emph{mesh-native, graph-based, and neural-field methods}, which are better aligned with FE discretizations and spatiotemporal field prediction. \citep{li2025new} provide one of the clearest crash-specific demonstrations of this idea by representing a B-pillar simulation as a graph sequence and combining graph coarsening with temporal recurrence. Their results show that architectural choices aimed at long-range message passing and rollout stability materially improve both accuracy and computational efficiency, which is especially relevant in crash problems where spatial resolution and temporal depth are both large. \citep{le2025comparing} make a complementary contribution by directly comparing a traditional surrogate approach with neural fields on scarce crash simulation data. This comparison is valuable because it reframes the question from ``can ML replace FE?'' to ``which representation is most appropriate under realistic data scarcity?'' Extending this direction to a larger structural setting, \citep{nabian2025automotive} present BIW-scale crash surrogate modeling in the PhysicsNeMo framework, comparing MeshGraphNet- and transformer-style architectures on a substantial crash dataset and showing large computational savings, albeit still with a remaining gap to full-FE fidelity. Collectively, these works indicate that mesh-aware learning is currently the most promising route for mesh-resolved spatiotemporal crash prediction.

A third family focuses on \emph{solver-integrated and hybrid methods}. Rather than predicting outputs entirely outside the FE loop, these methods insert learned components into the computational mechanics pipeline itself. \citep{thel2024introducing} introduce Finite Element Method Integrated Networks (FEMIN), where large regions of the crash mesh are replaced during runtime by neural surrogates that exchange interface quantities with the remaining FE model. The follow-up study of \citep{thel2025accelerating} compares force-predicting and kinematics-predicting FEMIN variants, showing that different integration strategies have different scaling behavior and that kinematics-driven coupling is especially attractive for larger load cases. Related ideas also appear at smaller scales: \citep{andre2023neural} model mechanical joints with feedforward neural networks embedded in an explicit crash solver, while \citep{pulikkathodi2025nonintrusive} use deep local models within a nonintrusive local/global coupling strategy for spot-welded structures under impact. These papers are particularly relevant for engineering deployment because they preserve strong compatibility with existing FE workflows, but they also tend to require deeper solver integration and are often harder to reproduce across datasets and software stacks.

Recent work also broadens the field beyond forward approximation toward \emph{benchmarking, interpretability, and research infrastructure}. \citep{rodriguez2025mechbench} propose MECHBench, a set of black-box optimization benchmarks rooted in structural mechanics and crashworthiness scenarios, directly addressing the lack of standardized application-oriented evaluation problems. At the same time, \citep{mathieu2026explainable} argue that explainable AI is necessary to turn crash-prediction models into engineering tools that improve system understanding rather than merely accelerate inference. This is an important shift: in safety-critical virtual development, accurate predictions alone are often insufficient unless engineers can also understand why a model prioritizes certain regions, components, or design variables. Together with \citep{liu2025intelligent}, these papers indicate that the next stage of the field will likely combine prediction quality with better human interpretability and stronger evaluation protocols.

Overall, the literature reveals three persistent gaps. First, reproducibility remains limited: many studies rely on restricted industrial datasets or narrowly defined component case studies, making cross-paper comparison difficult. Second, evaluation is fragmented across scalar metrics, component-level tasks, and differing solver contexts, so it is often unclear whether gains come from better representations, easier datasets, or weaker baselines. Third, only a small fraction of the literature jointly addresses spatiotemporal field prediction, interpretability, and standardized benchmarking. Our work is motivated by precisely these gaps. We introduce a new crash-oriented model and dataset with an emphasis on reproducibility, mesh-resolved spatiotemporal prediction, interpretable analysis, and standardized baselines spanning classical surrogates and modern mesh-aware methods.

\section{Limitations and Future Work}
\label{sec:limitations_future_work}

Although \textsc{CarCrashNet} advances open data-driven crash simulation, several
limitations remain. First, the current full-vehicle campaigns focus on three
baseline vehicles under frontal rigid-wall impact. This provides a controlled
benchmark, but does not yet cover offset, side, rear, oblique, rollover, or
pedestrian-safety scenarios. Expanding the dataset to additional impact modes
and boundary conditions is an important direction for future work.

Second, our validation primarily emphasizes global response quantities, including
force, energy, crash-pulse scale, and deformation trends. These quantities are
appropriate for dataset generation, but local acceleration channels remain more
solver-sensitive. Future work should include more systematic validation of local
kinematics, intrusion, accelerometer histories, and occupant-relevant metrics.

Third, the current machine-learning evaluation still relies partly on global
mean nodal errors such as RMSE and MAE. While these metrics are useful for
standardized comparison, they can be dominated by large nearly static regions of
the vehicle and may underemphasize localized crash-critical deformation. In
future benchmark releases, we will place greater emphasis on crash-aware metrics,
including relative displacement error, mass- or area-weighted displacement error,
crash-zone error, front-rail, bumper, and subframe component errors, final-frame
error, and high-percentile nodal errors such as the 95th-percentile displacement
error. Relative position error should be interpreted with caution, since it can
appear artificially small when normalized by the undeformed vehicle coordinate
scale.

Finally, while the dataset provides multi-modal outputs, the current benchmarks
cover only a subset of possible learning tasks. Future extensions will include
long-horizon rollout prediction, uncertainty quantification, inverse design,
failure detection, energy-consistent learning, and transfer across vehicle
classes and mesh resolutions.

\clearpage

\section{Solver Validation Against Experimental References and Ansys LS-DYNA}
\label{sec:solver_validation}

\paragraph{Motivation for OpenRadioss validation.}
OpenRadioss~\citep{openradioss_github} is a relatively recent open-source release of Altair Radioss~\citep{altair_radioss}:
Altair announced the open-source solver on September~8,~2022, with the goal of
building a broader community around industrial dynamic finite-element simulation. In contrast to older open-source solvers in
other domains, such as OpenFOAM in computational fluid dynamics, OpenRadioss is
still developing its public validation ecosystem and community benchmarks for
automotive crash simulation. Public crash validation studies are therefore
important not only for assessing solver credibility, but also for identifying
solver-sensitive quantities, documenting reproducible workflows, and supporting
broader community adoption. In this work, we treat the Toyota Yaris comparison
against published experimental references and Ansys LS-DYNA as an initial step
toward open verification and validation for data-driven crash simulation.

\subsection{Toyota Yaris Experimental and LS-DYNA Comparison}
\label{sec:yaris_solver_validation}

\paragraph{Validation protocol.}
We compare the OpenRadioss Toyota Yaris baselines against two external
references: published validation documentation from the Center for Collision
Safety and Analysis (CCSA) and the National Highway Traffic Safety
Administration (NHTSA), and an independently run Ansys LS-DYNA simulation.
The CCSA coarse and detailed Yaris validation reports use the same full-frontal
rigid-wall condition considered here, with NHTSA tests 5677 and 6221 providing
physical-test references at \SI{56.3}{\kilo\metre\per\hour} and
\SI{56.2}{\kilo\metre\per\hour}, respectively
\citep{ccsa_yaris_coarse_validation_2016,ccsa_yaris_detailed_validation_2016}.
The National Crash Analysis Center (NCAC) development and extended-validation
summaries report a maximum wall-force scale of approximately
\SI{550}{\kilo\newton}, an impact duration of approximately
\SI{100}{\milli\second}, and a passing total-wall-force validation metric for
LS-DYNA relative to the physical tests
\citep{ncac_yaris_development_validation_2011,ncac_yaris_extended_validation_2012}.

For wall-force quantities, we use direct wall-output histories rather than the
momentum-derivative proxy. OpenRadioss force is taken from the dominant
\texttt{/TH/RWALL} wall-normal component, LS-DYNA force is taken from the
\texttt{rwforc} impact-direction component, and both signals are filtered using
the standard Channel Frequency Class 60 (CFC60) crash-test low-pass filter
before scalar peak and duration extraction.

\paragraph{Scalar response agreement.}
Table~\ref{tab:solver_validation_global} summarizes the coarse OpenRadioss,
detailed OpenRadioss, and LS-DYNA responses against the published CCSA/NHTSA
validation context. All solver runs reproduce the
\SI{56.3}{\kilo\metre\per\hour} impact-speed regime. The detailed OpenRadioss
model is closer to LS-DYNA in wall-force duration and internal energy than the
coarse OpenRadioss model, while contact energy remains the most
solver-sensitive scalar.

\begin{table*}[!htbp]
  \centering
  \caption{
    Yaris full-frontal validation summary. Wall force is the CFC60-filtered
    direct wall-output force. OpenRadioss
    uses the dominant \texttt{TH-RWALL} wall-normal component; LS-DYNA uses the
    \texttt{rwforc} impact-direction component. The reference column summarizes
    the published CCSA/NHTSA validation context.
  }
  \label{tab:solver_validation_global}
  \scriptsize
  \setlength{\tabcolsep}{3.5pt}
  \renewcommand{\arraystretch}{1.08}
  \begin{tabular*}{\textwidth}{@{\extracolsep{\fill}} l r r r p{3.5cm} @{}}
    \toprule
    \textbf{Metric} &
    \shortstack{\textbf{OpenRadioss}\\\textbf{coarse}} &
    \shortstack{\textbf{OpenRadioss}\\\textbf{detailed}} &
    \shortstack{\textbf{LS-DYNA}\\\textbf{coarse}} &
    \shortstack[l]{\textbf{Published reference}\\\textbf{context}} \\
    \midrule
    Initial speed [km/h] & 56.32 & 56.32 & 56.32 &
    NHTSA 5677/6221: 56.3/56.2 \\
    CFC60 peak wall force [kN] & 654.8 & 630.3 & 588.1 &
    About 550 kN maximum wall-force scale \\
    CFC60 wall-force duration [ms] & 68.2 & 80.4 & 78.4 &
    About 100 ms impact duration \\
    Peak internal energy [kJ] & 153.3 & 143.4 & 142.7 &
    Energy balance reported in source validation \\
    Peak contact energy [kJ] & 18.9 & 14.1 & 8.8 &
    Solver-sensitive; no direct test scalar \\
    Final energy error [\%] & 10.2 & -1.0 & -2.8 &
    No direct test scalar \\
    \bottomrule
  \end{tabular*}
\end{table*}
\FloatBarrier

\begin{figure}[!htbp]
    \centering
    \includegraphics[width=0.92\textwidth]{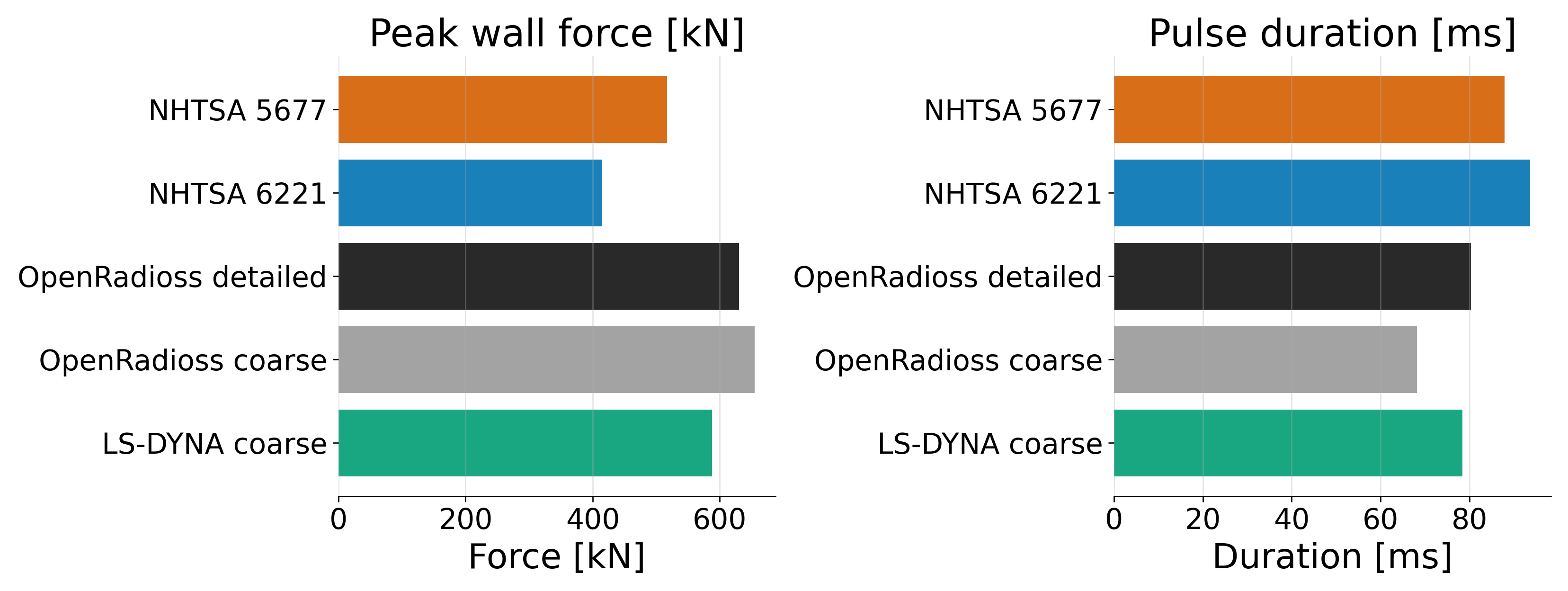}
    \caption{Scalar wall-force validation summary. NHTSA 5677 and 6221 values are digitized from the CCSA validation
    material; solver values are extracted from CFC60-filtered direct
    wall-output histories.}
    \label{fig:wall_force_scalar_summary}
\end{figure}

Fig.~\ref{fig:wall_force_scalar_summary} makes the physical-test force and
duration scale explicit by comparing the solver scalars against the digitized
NHTSA 5677 and 6221 wall-force references from the CCSA validation material.

\FloatBarrier

\begin{figure}[h!]
    \centering
    \begin{subfigure}[t]{0.85\textwidth}
        \centering
        \includegraphics[width=\textwidth,height=0.22\textheight,keepaspectratio]{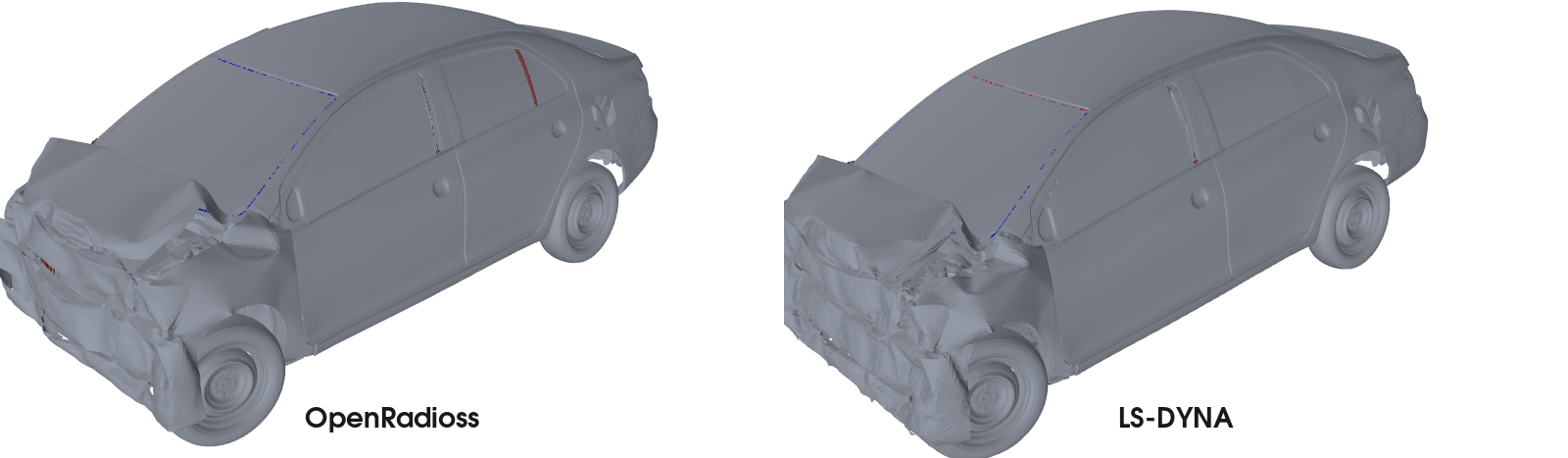}
        \caption{Isometric comparison between OpenRadioss and LS-DYNA.}
        \label{fig:iso_comparison}
    \end{subfigure}

    \begin{subfigure}[t]{0.85\textwidth}
        \centering
        \includegraphics[width=\textwidth,height=0.22\textheight,keepaspectratio]{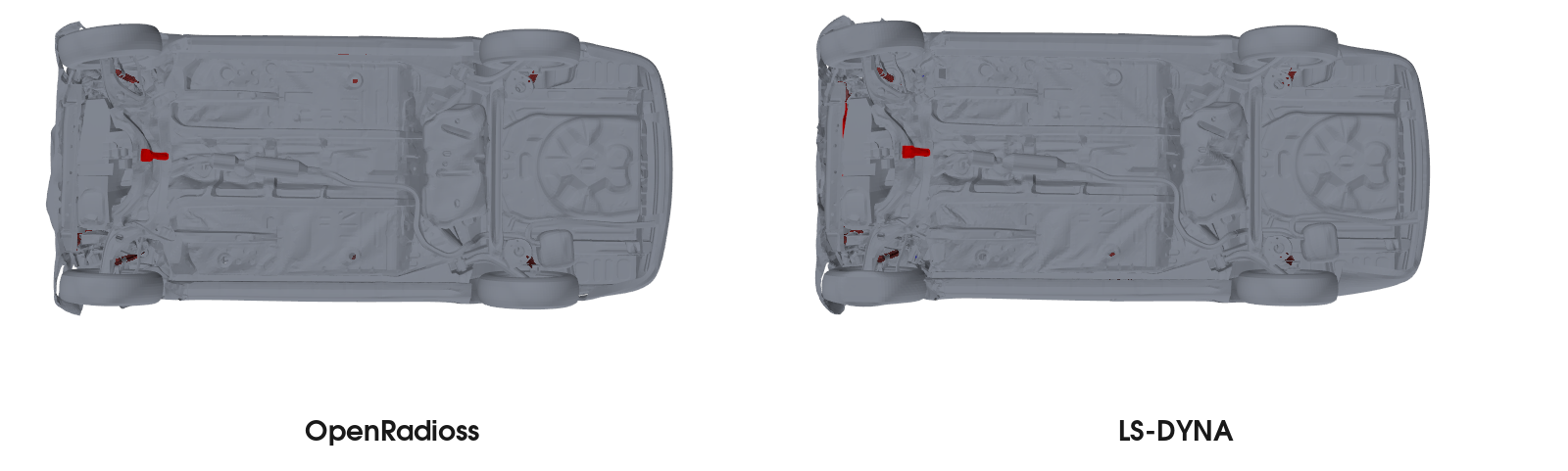}
        \caption{Underbody comparison after frontal impact.}
        \label{fig:underbody_comparison}
    \end{subfigure}

    \begin{subfigure}[t]{0.85\textwidth}
        \centering
        \includegraphics[width=\textwidth,height=0.22\textheight,keepaspectratio]{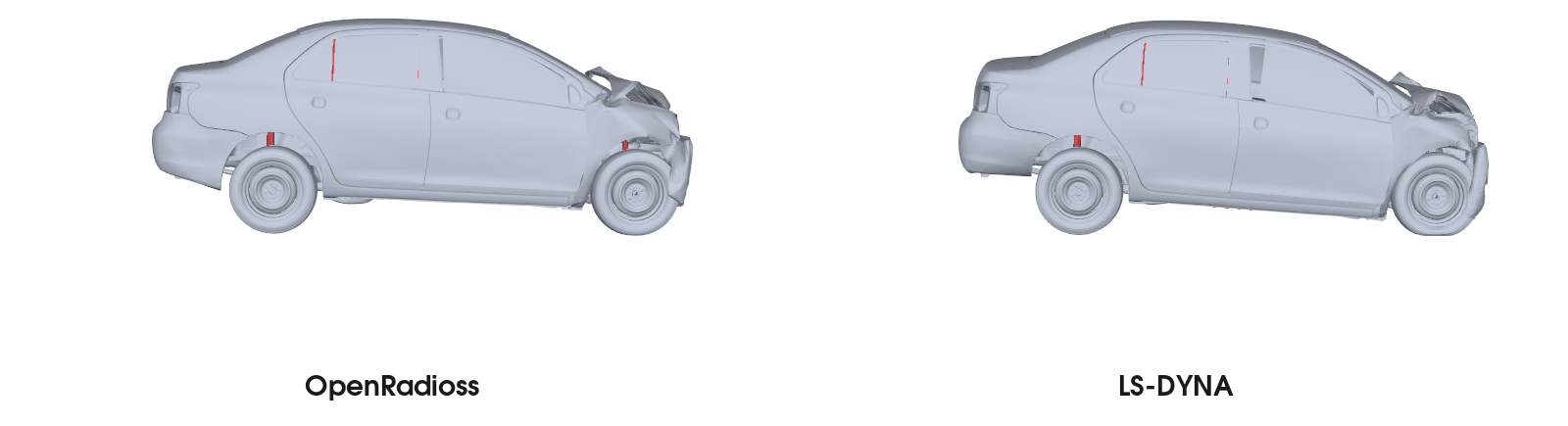}
        \caption{Side-view comparison of the deformed Yaris model.}
        \label{fig:side_comparison}
    \end{subfigure}

    \caption{Qualitative comparison of post-impact deformation fields between
    OpenRadioss and LS-DYNA for the same Yaris full-frontal crash configuration.
    Across isometric, underbody, and side views, the two solvers produce broadly
    consistent front-end crush morphology and global deformation patterns.}
    \label{fig:deformation_comparison_views}
\end{figure}

\paragraph{Deformation morphology.}
The qualitative comparison in Fig.~\ref{fig:deformation_comparison_views}
shows that OpenRadioss and LS-DYNA produce broadly consistent front-end crush
morphology and global deformation patterns. This supports using the OpenRadioss
workflow for global-response dataset generation.

\FloatBarrier

\begin{figure}[h!]
    \centering
    \begin{subfigure}[t]{0.98\textwidth}
        \centering
        \includegraphics[width=\textwidth]{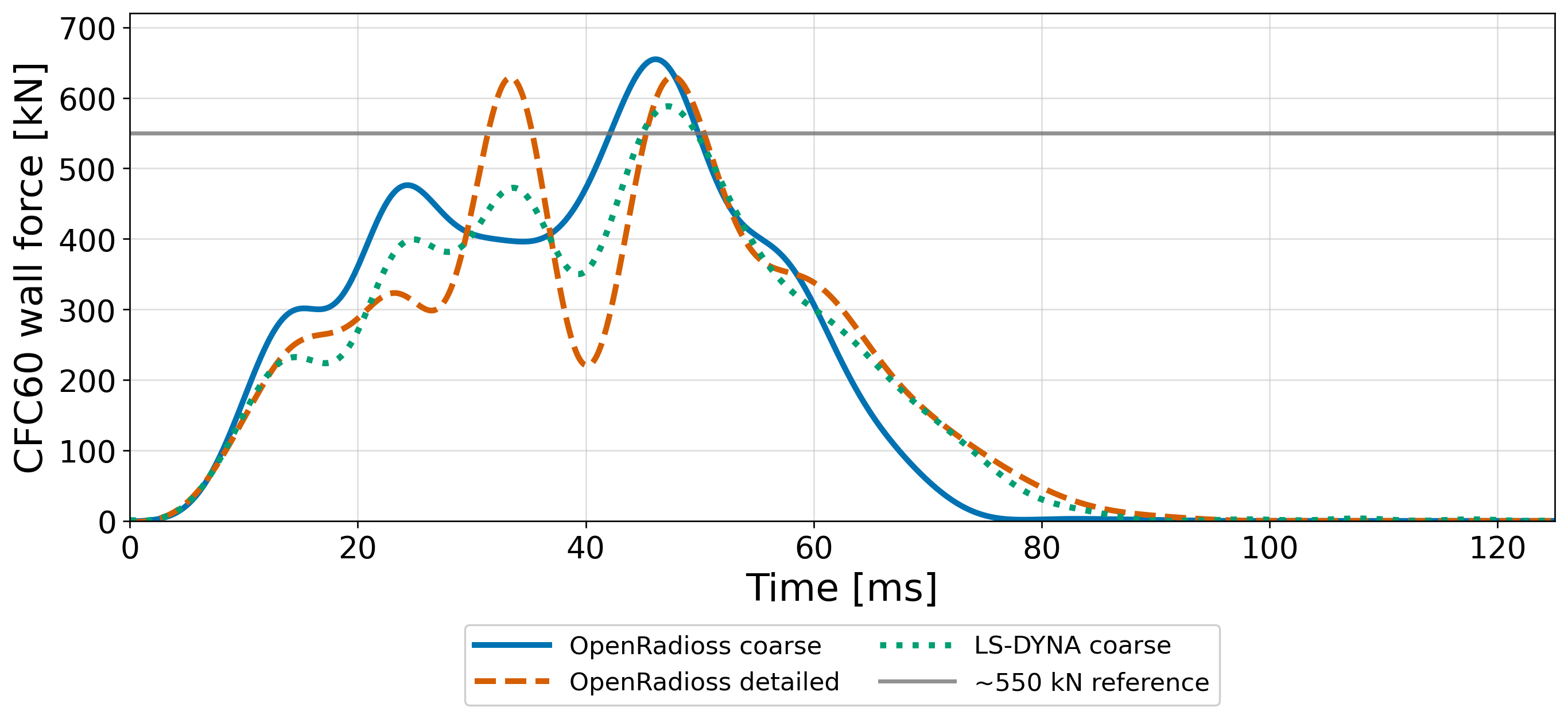}
        \caption{Comparison of CFC60-filtered direct wall-force histories.
        OpenRadioss uses the dominant \texttt{TH-RWALL} wall-normal component,
        LS-DYNA uses the \texttt{rwforc} impact-direction component, and the
        horizontal line marks the approximate NCAC-reported maximum wall-force
        scale.}
        \label{fig:wall_force_overlay}
    \end{subfigure}

    \vspace{0.6em}
    \begin{subfigure}[t]{0.98\textwidth}
        \centering
        \includegraphics[width=0.96\textwidth]{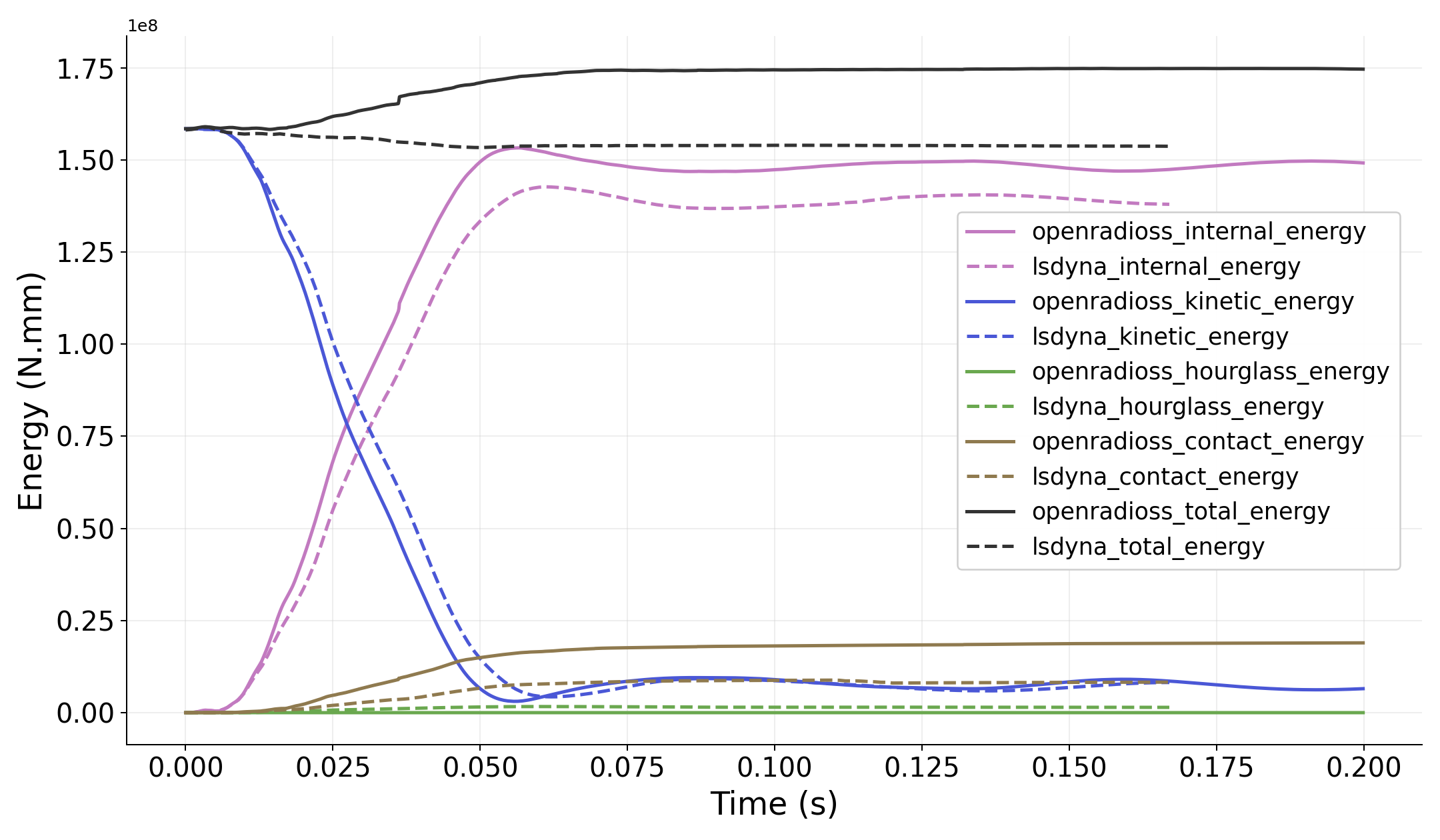}
        \caption{Comparison of global energy time histories between
        OpenRadioss and LS-DYNA, including internal, kinetic, hourglass,
        contact, and total energy components.}
        \label{fig:energy_overlay}
    \end{subfigure}
    \caption{Validation comparison of wall-force and energy responses for the
    same Yaris full-frontal impact configuration.}
    \label{fig:solver_comparison}
\end{figure}

\paragraph{Time-history agreement.}
Figure~\ref{fig:solver_comparison} compares the solver wall-force and energy
time histories for the same full-frontal impact configuration. The direct
wall-force histories in Fig.~\ref{fig:wall_force_overlay} show similar pulse
timing and decay scale across OpenRadioss and LS-DYNA, with OpenRadioss higher
in peak force relative to the NCAC-reported maximum-force scale. The energy
histories in Fig.~\ref{fig:energy_overlay} show comparable
kinetic-to-internal energy conversion trends, with the main discrepancy in
contact and total energy. We therefore use the OpenRadioss Yaris runs for
global deformation, force, and energy learning, while treating local
acceleration and contact-sensitive quantities as solver-sensitive diagnostics.

\FloatBarrier

\subsection{Timestep Sensitivity and Production Baseline}
\label{sec:timestep_sensitivity}

To evaluate the numerical sensitivity of our simulation, we ran a separate four-case
timestep sweep using the same coarse Yaris model and full-frontal rigid-wall
boundary condition. The sweep cases were \texttt{dt2ms\_off}, \texttt{dt2ms\_q25},
\texttt{dt2ms\_q50}, and \texttt{dt2ms\_q100}. The \texttt{q25}, \texttt{q50}, and
\texttt{q100} labels denote $25\%$, $50\%$, and $100\%$ of the production timestep
target, respectively. The \texttt{off} case used \texttt{/DT} without a fixed nodal
constant target. These runs do not change the vehicle geometry, materials, impact
condition, or DoE design variables; they only change the explicit timestep control.
We therefore treat them as numerical-sensitivity runs, not as additional dataset
baselines.

Table~\ref{tab:timestep_sweep_full} summarizes the full sweep against the Ansys
LS-DYNA coarse reference. The most conservative scalar setting, \texttt{dt2ms\_q25},
is the slowest run, while \texttt{dt2ms\_q50} gives the best combined score once
wall-force history, energy histories, and local acceleration peaks are considered,
with substantially lower runtime. In contrast, \texttt{dt2ms\_q100} gives a
superficially closer wall-force peak but degrades the internal-energy response and
the overall validation score.

\begin{table}[h!]
  \centering
  \footnotesize
  \caption{Timestep-sensitivity sweep for the coarse Toyota Yaris model. Relative differences are reported with respect to the Ansys LS-DYNA coarse reference. Lower score is better.}
  \label{tab:timestep_sweep_full}
  \setlength{\tabcolsep}{4pt}
  \renewcommand{\arraystretch}{1.08}
  \begin{tabular*}{\textwidth}{@{\extracolsep{\fill}} l c c c c c @{}}
    \toprule
    \textbf{Case} &
    \textbf{$\Delta t$ [$\mu$s]} &
    \textbf{Runtime [min]} &
    \textbf{Wall peak [kN]} &
    \textbf{IE peak [kJ]} &
    \textbf{Score} \\
    \midrule
    \texttt{dt2ms\_off}  & none  & 262.1 & 664.7 {\scriptsize $(+13.5\%)$} & 156.7 {\scriptsize $(+9.8\%)$}  & 43.1 \\
    \texttt{dt2ms\_q25}  & 0.250 & 447.9 & 670.5 {\scriptsize $(+14.5\%)$} & 152.7 {\scriptsize $(+7.0\%)$}  & 44.7 \\
    \texttt{dt2ms\_q50}  & 0.500 & 96.6  & 656.9 {\scriptsize $(+12.2\%)$} & 151.8 {\scriptsize $(+6.4\%)$}  & \textbf{41.1} \\
    \texttt{dt2ms\_q100} & 1.001 & 48.4  & 641.3 {\scriptsize $(+9.5\%)$}  & 181.1 {\scriptsize $(+26.9\%)$} & 62.9 \\
    \bottomrule
  \end{tabular*}
\end{table}

\clearpage

\section{Vehicle-Scale Crash Dataset and Baseline FE Models}
\label{app:vehicle_baseline_models}

\begin{figure}[h!]
    \centering
    \includegraphics[width=\linewidth]{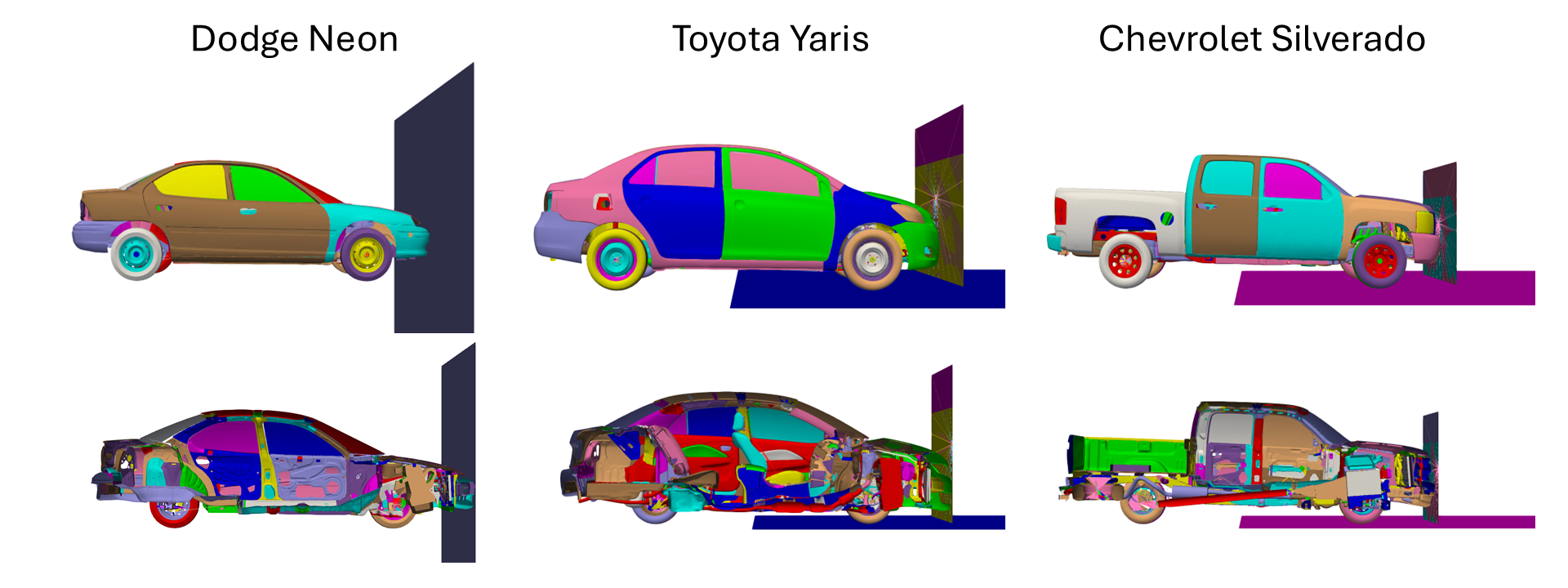}
    \caption{
      Baseline vehicle models used for the dataset generation in
      \textsc{CarCrashNet}. From left to right: Dodge Neon, Toyota Yaris, and
      Chevrolet Silverado. The top row shows exterior geometry and the bottom
      row shows cutaway views exposing structural and cabin components. The
      three models span different vehicle classes, mesh resolutions, and
      frontal load-path layouts.
    }
    \label{fig:carcrashnet_baselines}
\end{figure}

The vehicle-scale portion of \textsc{CarCrashNet} is built on three
full-vehicle finite-element baselines: a Toyota Yaris passenger car, a
Dodge Neon passenger car, and a detailed Chevrolet Silverado pickup truck (Fig.~\ref{fig:carcrashnet_baselines}).
Each baseline is used as a fixed reference model. The dataset-generation
campaigns modify the impact speed and the selected shell-thickness groups; they do
not recalibrate material parameters, alter the contact formulation, or simplify
the original part topology. This is important for
scientific machine learning because it preserves realistic structural
heterogeneity while keeping the design space interpretable and reproducible.

\subsection{Shared Campaign Design Space}
\label{app:vehicle_shared_design_space}

All three vehicle-scale campaigns use the same abstract design vector,
\begin{equation}
  \boldsymbol{\xi}
  =
  \left[
    v,\,
    s_{\mathrm{front}},\,
    s_{\mathrm{rail}}
  \right],
  \label{eq:app_vehicle_design_vector}
\end{equation}
where $v$ is the impact velocity, $s_{\mathrm{front}}$ scales the selected
front-support shell thicknesses, and $s_{\mathrm{rail}}$ scales the selected
lower-rail or subframe shell thicknesses. Across all three campaigns,
$v \in [50,64]\,\si{\kilo\metre\per\hour}$,
$s_{\mathrm{front}} \in [0.9,1.1]$, and
$s_{\mathrm{rail}} \in [0.9,1.1]$.

For Toyota Yaris and Chevrolet Silverado, the interior design points are
generated by Latin-hypercube sampling~\citep{mckay1979lhs} after reserving deterministic anchor
cases at the baseline, one-factor extrema, and design-space corners. For Dodge
Neon, the first 75 cases establish the same anchor-backed pilot structure,
while subsequent batches use greedy maximin continuation~\citep{johnson1990minimax}. If $\mathcal{X}_k$ is
the accumulated normalized design set and $\mathcal{C}$ is a candidate pool, the
next continuation point is
\begin{equation}
  \mathbf{x}_{k+1}
  =
  \arg\max_{\mathbf{x}\in\mathcal{C}}
  \min_{\mathbf{z}\in\mathcal{X}_k}
  \left\lVert \mathbf{x}-\mathbf{z}\right\rVert_2 .
  \label{eq:app_vehicle_maximin}
\end{equation}
This expands training coverage without changing the fixed anchor and held-out
test structure.

\subsection{Toyota Yaris Coarse Baseline}
\label{app:vehicle_baseline_yaris}

\paragraph{Baseline and response.}
The Yaris campaign uses the CCSA coarse 2010 Toyota Yaris model, a
heterogeneous shell-dominated vehicle deck with \num{919} parts,
\num{393164} nodes, and \num{358457} shell elements. The baseline
full-frontal rigid-wall run reaches an initial speed of
\SI{56.32}{\kilo\metre\per\hour}, absorbs $95.89\%$ of its initial kinetic
energy, and produces peak internal and contact energies of
\SI{153.27}{\kilo\joule} and \SI{18.94}{\kilo\joule}. Using the same
CFC60-filtered direct wall-output extraction as
Table~\ref{tab:solver_validation_global}, the coarse OpenRadioss run gives a
\SI{654.8}{\kilo\newton} peak wall force and a
\SI{68.2}{\milli\second} force-pulse duration.

\paragraph{Campaign role.}
The 500-case Yaris campaign contains 15 deterministic anchors and 485
Latin-hypercube samples. It scales 27 front-support shell sections and 10
lower-rail/subframe sections, making it the main high-throughput
passenger-car benchmark for deformation, global force, and energy learning.

\subsection{Dodge Neon Full-Frontal Baseline}
\label{app:vehicle_baseline_neon}

\paragraph{Baseline and response.}
The Neon campaign uses the NCAC Dodge Neon V04i full-frontal model in native
Radioss form. The vehicle has \num{336} parts, \num{283859} nodes, and
\num{267786} shell elements, with a finer median shell-edge length than the
coarse Yaris. The OpenRadioss reference run reaches
\SI{56.16}{\kilo\metre\per\hour}, absorbs $95.39\%$ of its initial kinetic
energy, and gives peak internal, contact, and hourglass energies of
\SI{132.84}{\kilo\joule}, \SI{1.02}{\kilo\joule}, and
\SI{9.22}{\kilo\joule}. Neon histories are therefore used mainly for global
energy, mass, and termination diagnostics.

\paragraph{Campaign role.}
Neon uses the same abstract design variables as Yaris but applies them to a
different mesh, part hierarchy, material inventory, and frontal load path. The
campaign varies 12 front-support and 15 lower-rail/subframe shell properties;
after a 75-case anchor-backed pilot, continuation batches add train-only
maximin samples. This makes Neon the cross-vehicle passenger-car test case.

\subsection{Chevrolet Silverado Detailed Baseline}
\label{app:vehicle_baseline_silverado}

\paragraph{Baseline and response.}
The Silverado campaign uses the detailed 2007 Chevrolet Silverado CCSA V3e
model. It is the largest baseline in the corpus, with \num{719} parts,
\num{979490} nodes, \num{907067} shell elements, and \num{53281} solid
elements, and it introduces a pickup-truck frame architecture rather than a
passenger-car unibody. The structural-campaign baseline reaches
\SI{57.34}{\kilo\metre\per\hour}, absorbs $98.49\%$ of its initial kinetic
energy, and produces peak internal, contact, and hourglass energies of
\SI{274.56}{\kilo\joule}, \SI{16.32}{\kilo\joule}, and
\SI{2.91}{\kilo\joule}.

\subsection{Scientific Role of the Three Campaigns}
\label{app:vehicle_campaigns:role}

The three campaigns are complementary rather than redundant. Neon and Yaris
provide two passenger-car baselines with a shared structural-loading coordinate
system but different mesh statistics and load-path layouts. Yaris provides the
largest completed vehicle-scale campaign, while Neon enables explicit
cross-vehicle transfer tests without changing the abstract DoE variables.
Silverado extends the same design-space logic to a larger pickup-truck model,
increasing both geometric complexity and solver cost.

This organization gives the dataset three useful properties for scientific
machine learning. First, the design spaces are low-dimensional and physically
interpretable, which makes extrapolation and failure modes easier to diagnose.
Second, every campaign retains both field trajectories and scalar histories, so
models can be evaluated on deformation fields, energy/force consistency, and
reduced crashworthiness quantities. Third, the shared abstraction across
vehicles enables controlled studies of within-vehicle interpolation,
cross-vehicle transfer, and scaling from passenger-car meshes to a detailed
truck mesh.

\subsection{Cross-Vehicle Mesh Comparison}
\label{sec:vehicle_mesh_comparison}

To quantify the diversity of the vehicle-scale portion of
\textsc{CarCrashNet}, we compare the undeformed finite-element meshes of the
three full-vehicle baselines used for dataset generation. The comparison is computed
from the source vehicle decks prior to impact. Rigid-wall and
campaign-specific loading files are excluded, so the reported values describe
the structural vehicle models rather than the boundary-condition geometry.

This comparison is useful for two reasons. First, it verifies that the Neon and
Silverado campaigns are not merely additional samples from the Yaris mesh
distribution. Second, it exposes the numerical scale seen by machine-learning
models: node count, element count, part granularity, shell resolution, and
time-resolved VTKHDF storage size differ substantially across the three
vehicles.

\paragraph{Global finite-element complexity.}
Table~\ref{tab:vehicle_mesh_global} summarizes the global mesh statistics. Total elements include shell, solid, beam, and auxiliary element
records present in the source vehicle deck. The Silverado model is the largest
by a wide margin, with \num{979490} nodal records and \num{963659} total
element records. It contains \num{907067} shell elements, \num{53281} solid
elements, and a $7.404$~GB baseline VTKHDF trajectory, making it roughly
$4.45\times$ larger than the Yaris VTKHDF artifact and $5.69\times$ larger
than the Neon artifact.

\begin{table}[h!]
  \centering
  \small
  \caption{
    Global finite-element model complexity for the three vehicle baselines.
    Rigid-wall and loading-only include files are excluded. The VTKHDF size is
    measured from a representative baseline trajectory for each vehicle.
  }
  \label{tab:vehicle_mesh_global}
  \setlength{\tabcolsep}{4pt}
  \resizebox{\textwidth}{!}{%
  \begin{tabular}{lrrrrrrrrr}
    \toprule
    \textbf{Vehicle} &
    \textbf{Deck [MB]} &
    \textbf{VTKHDF [GB]} &
    \textbf{Nodes} &
    \textbf{Total elems.} &
    \textbf{Shells} &
    \textbf{Solids} &
    \textbf{Beams} &
    \textbf{Parts} &
    \textbf{Material cards} \\
    \midrule
    Toyota Yaris  &
    42.85 & 1.665 &
    \num{393164} & \num{378550} & \num{358457} &
    \num{15234} & \num{4685} & \num{919} & \num{919} \\
    Dodge Neon &
    30.74 & 1.301 &
    \num{283859} & \num{271111} & \num{267786} &
    \num{2852} & \num{122} & \num{336} & \num{336} \\
    Chevrolet Silverado &
    105.76 & 7.404 &
    \num{979490} & \num{963659} & \num{907067} &
    \num{53281} & \num{3113} & \num{719} & \num{719} \\
    \bottomrule
  \end{tabular}%
  }
\end{table}

\paragraph{Connection and constraint inventory.}
Crash models contain many non-continuum definitions that are important for
load transfer but are not captured by shell, solid, and beam counts alone.
Table~\ref{tab:vehicle_connection_inventory} audits the main connector and
constraint families in the same source decks. Nodal rigid bodies represent
rigid kinematic couplings between a reference node and a node set. Extra-node
constraints add nodes to an existing rigid body. Joint constraints represent
idealized revolute, spherical, cylindrical, or translational joints.
Rigid-body merge records couple two or more rigid bodies. Spot-weld entries
represent local welded connector definitions. These counts are source-deck
records, not material cards; for example, a single spot-weld material model may
be reused by thousands of connector definitions.

\begin{table}[h!]
\centering
\footnotesize
\caption{Connector and constraint inventory for the three baseline vehicle models used in \textsc{CarCrashNet}.}
\label{tab:vehicle_connection_inventory}
\setlength{\tabcolsep}{4pt}
\renewcommand{\arraystretch}{1.08}
\begin{tabular*}{\textwidth}{@{\extracolsep{\fill}} l c c c c c @{}}
\toprule
\textbf{Vehicle} &
\shortstack[c]{\textbf{Nodal rigid}\\ \textbf{bodies}} &
\shortstack[c]{\textbf{Extra-node}\\ \textbf{constraints}} &
\shortstack[c]{\textbf{Joint}\\ \textbf{constraints}} &
\shortstack[c]{\textbf{Rigid-body}\\ \textbf{merges}} &
\shortstack[c]{\textbf{Spot-weld}\\ \textbf{connectors}} \\
\midrule
Toyota Yaris       & 759 & 20 & 44 & 2  & 5656 \\
Dodge Neon       & 672 & 38 & 12 & 22 & 4197 \\
Chevrolet Silverado  & 696 & 56 & 40 & 4  & 7136 \\
\bottomrule
\end{tabular*}

\vspace{2pt}
\begin{minipage}{\textwidth}
\footnotesize
\textit{Note:} Spot-weld counts denote connector instances; for the Dodge Neon, this includes generalized weld-spot definitions.
\end{minipage}
\end{table}

\paragraph{Local shell resolution and vehicle envelope.}
The three models also differ in local discretization, not only in total size.
Table~\ref{tab:vehicle_mesh_resolution} reports the median shell-edge length,
median shell area, and axis-aligned vehicle envelope spans. Because the source
decks use different coordinate conventions, we report span magnitudes rather
than signed coordinate directions. Neon has fewer elements than Yaris but a
finer shell mesh, with a median shell-edge length of \SI{12.08}{\milli\metre}
compared with \SI{15.65}{\milli\metre} for Yaris. Silverado is both larger and
locally finer, with a median shell-edge length of \SI{10.43}{\milli\metre} and
a median shell area of \SI{106.39}{\milli\metre\squared}.

\begin{table}[h!]
\centering
\footnotesize
\caption{Local shell resolution and undeformed vehicle-envelope spans for the three baseline vehicle models used in \textsc{CarCrashNet}. Span values are reported as axis-aligned mesh extents and shown as magnitudes because the source decks do not share a common coordinate orientation.}
\label{tab:vehicle_mesh_resolution}
\setlength{\tabcolsep}{4pt}
\renewcommand{\arraystretch}{1.08}
\begin{tabular*}{\textwidth}{@{\extracolsep{\fill}} l c c c c c @{}}
\toprule
\textbf{Vehicle} &
\shortstack[c]{\textbf{Median edge}\\ \textbf{[mm]}} &
\shortstack[c]{\textbf{Median shell area}\\ \textbf{[mm$^2$]}} &
\shortstack[c]{\textbf{Length-like span}\\ \textbf{[mm]}} &
\shortstack[c]{\textbf{Width-like span}\\ \textbf{[mm]}} &
\shortstack[c]{\textbf{Height-like span}\\ \textbf{[mm]}} \\
\midrule
Toyota Yaris      & 15.65 & 235.23 & 4298.98 & 1964.82 & 1467.95 \\
Dodge Neon       & 12.08 & 138.92 & 4535.18 & 1721.64 & 1370.21 \\
Chevrolet Silverado  & 10.43 & 106.39 & 5846.05 & 2116.33 & 1919.53 \\
\bottomrule
\end{tabular*}
\end{table}

\paragraph{Implications for dataset design.}
The three vehicle baselines should be treated as distinct mesh distributions
rather than as replicas at different sample counts. Yaris provides the largest
completed passenger-car campaign and has the richest part granularity among the
two smaller vehicles. Neon has fewer parts and elements but a finer shell
discretization than Yaris, changing both graph topology and local geometric
statistics. Silverado is a different scale of problem: it has a longer vehicle
envelope, the finest median shell resolution, the largest solid-element
population, and substantially larger VTKHDF artifacts.

Solver time and postprocessing cost are also
controlled by explicit time-step size, contact activity, output frequency,
solid-element content, and VTKHDF I/O volume. Consequently, the three campaigns
are best viewed as complementary datasets for cross-vehicle generalization:
Yaris establishes a high-throughput coarse passenger-car distribution, Neon
tests transfer to a second passenger-car topology, and Silverado tests whether
the same workflow scales to a detailed truck model with much larger field
outputs.

Table~\ref{tab:vehicle_campaign_compute_cost} reports the solver cost of the vehicle-scale campaigns, totaling 91{,}648 core-hours across 825 simulations. 

\begin{table}[h!]
\centering
\scriptsize
\caption{Solver cost for the vehicle-scale \textsc{CarCrashNet} campaigns.}
\label{tab:vehicle_campaign_compute_cost}
\setlength{\tabcolsep}{4pt}
\renewcommand{\arraystretch}{1.05}
\begin{tabular*}{\textwidth}{@{\extracolsep{\fill}} l c c c c c @{}}
\toprule
\textbf{Vehicle} &
\textbf{Ranks} &
\textbf{Wall h/case} &
\textbf{Core-h/case} &
\textbf{Cases} &
\textbf{Total core-h} \\
\midrule
Toyota Yaris          & 64  & 1.42 & 90.8  & 500 & 45,440 \\
Dodge Neon            & 64  & 1.49 & 95.4  & 250 & 23,840 \\
Chevrolet Silverado   & 128 & 2.33 & 298.2 & 75  & 22,368 \\
\midrule
\textbf{Total} & -- & -- & -- & 825 & 91,648 \\
\bottomrule
\end{tabular*}

\end{table}

\begin{figure}[h!]
    \centering
    \includegraphics[width=\linewidth]{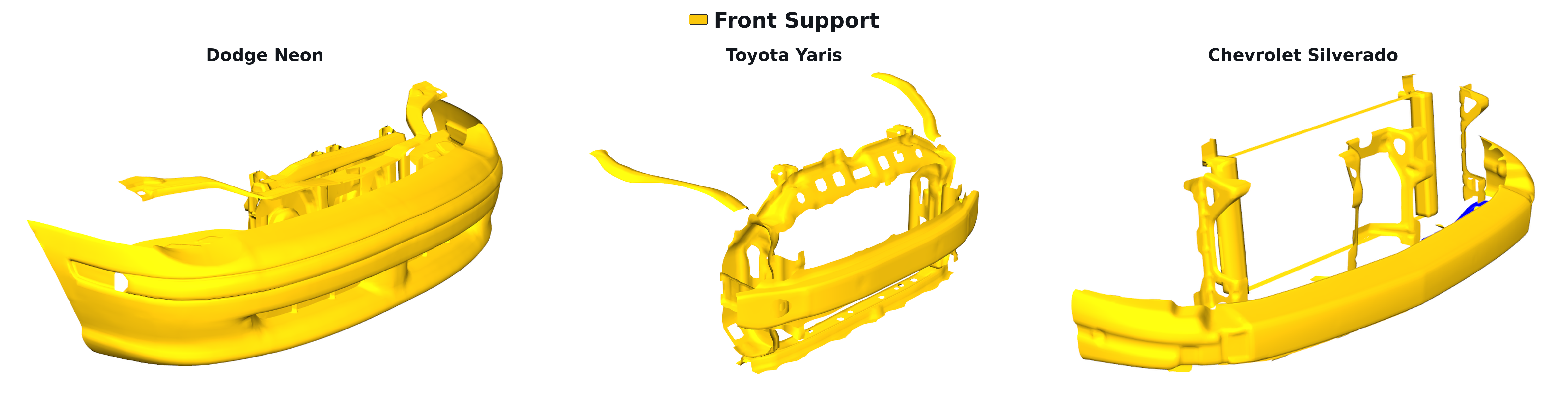}
    \caption{Front-support components whose thicknesses are varied as design parameters for vehicle-scale dataset generation in \textsc{CarCrashNet}. The highlighted structures differ substantially across the Dodge Neon, Toyota Yaris, and Chevrolet Silverado, reflecting the diversity of frontal crash architectures across vehicle classes. Because these components directly influence frontal load transfer, crush initiation, and energy absorption, varying their thickness expands the structural design space and improves dataset diversity for learning crash-safety-relevant responses.}
    \label{fig:vehicle_front_support_only}
\end{figure}

\begin{figure}[h!]
    \centering
    \includegraphics[width=\linewidth]{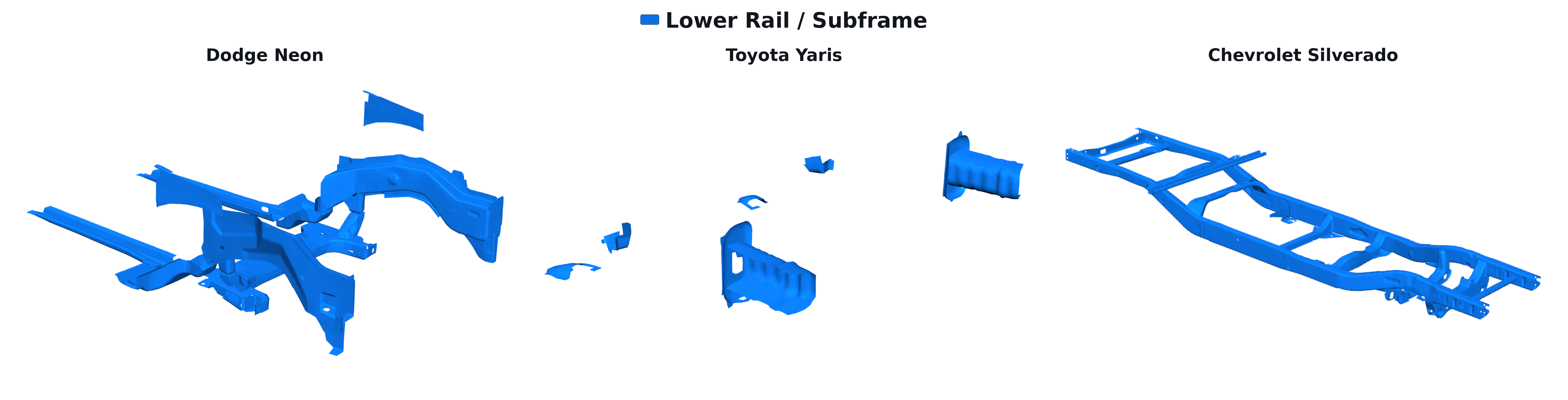}
    \caption{Lower-rail and subframe components whose thicknesses are varied as design parameters for vehicle-scale dataset generation in \textsc{CarCrashNet}. The highlighted structures exhibit strong geometric and topological variation across the Dodge Neon, Toyota Yaris, and Chevrolet Silverado, ranging from compact lower load paths to extended frame-like structures. Varying their thickness changes structural stiffness, load-path distribution, intrusion behavior, and global deformation modes, making these components central to the crash-safety design space.}
    \label{fig:vehicle_lower_rail_subframe_only}
\end{figure}

Figures~\ref{fig:vehicle_front_support_only} and
\ref{fig:vehicle_lower_rail_subframe_only} show the two structural regions whose
thicknesses are varied in the vehicle-scale design space. These components are
central to crashworthiness because they define the primary frontal load paths,
control crush initiation and progressive collapse, and strongly influence energy
absorption, intrusion, and deceleration during impact.

\begin{figure}[h!]
    \centering
    \includegraphics[
        width=\linewidth,
        height=0.95\textheight,
        keepaspectratio
    ]{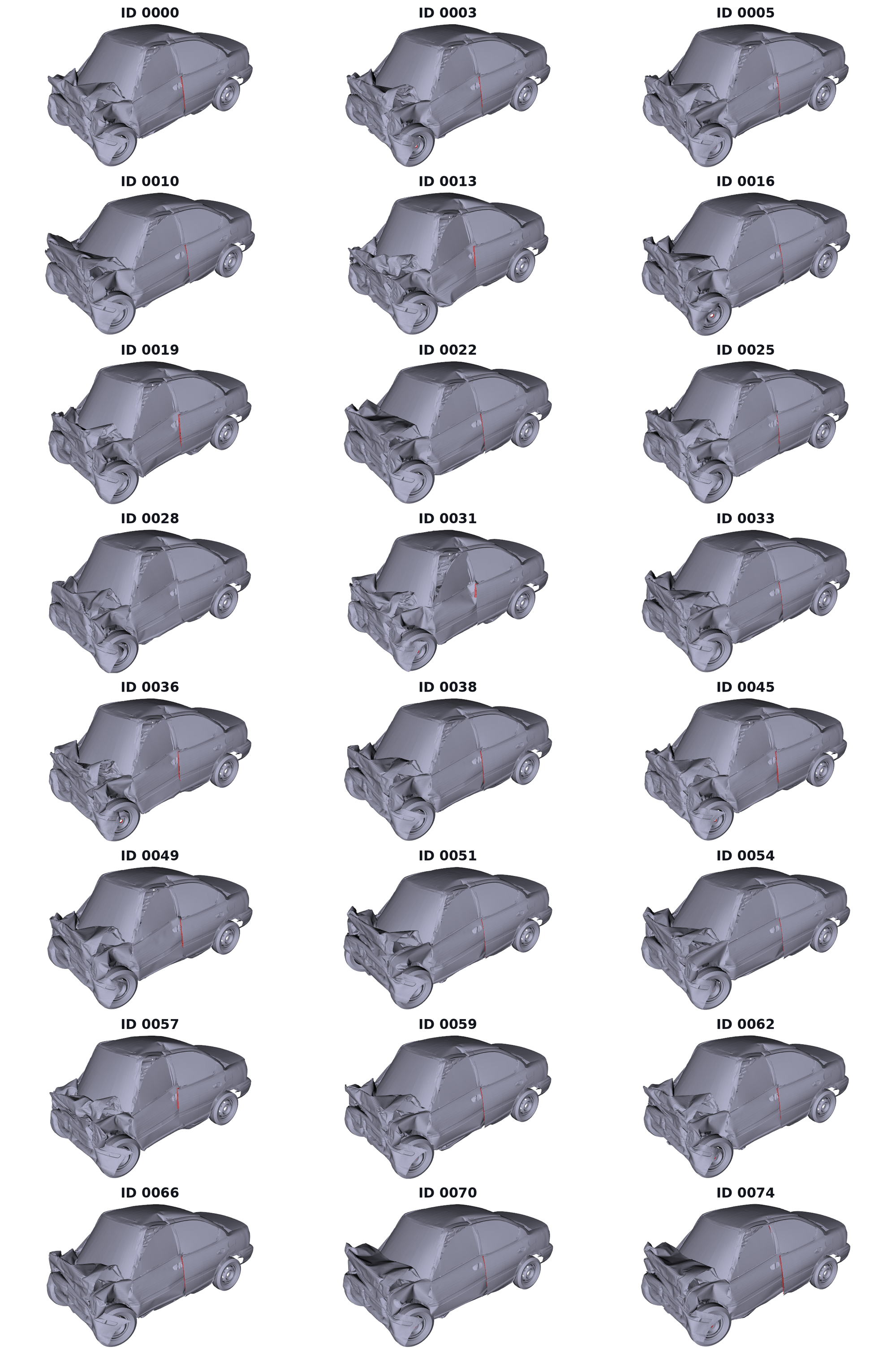}
    \caption{Representative deformed configurations from the Dodge Neon campaign shown in isometric view. The gallery demonstrates the range of full-vehicle crash outcomes captured for this platform in \textsc{CarCrashNet}.}
    \label{fig:neon_campaign_iso}
\end{figure}

\begin{figure}[h!]
    \centering
    \includegraphics[
        width=\linewidth,
        height=0.95\textheight,
        keepaspectratio
    ]{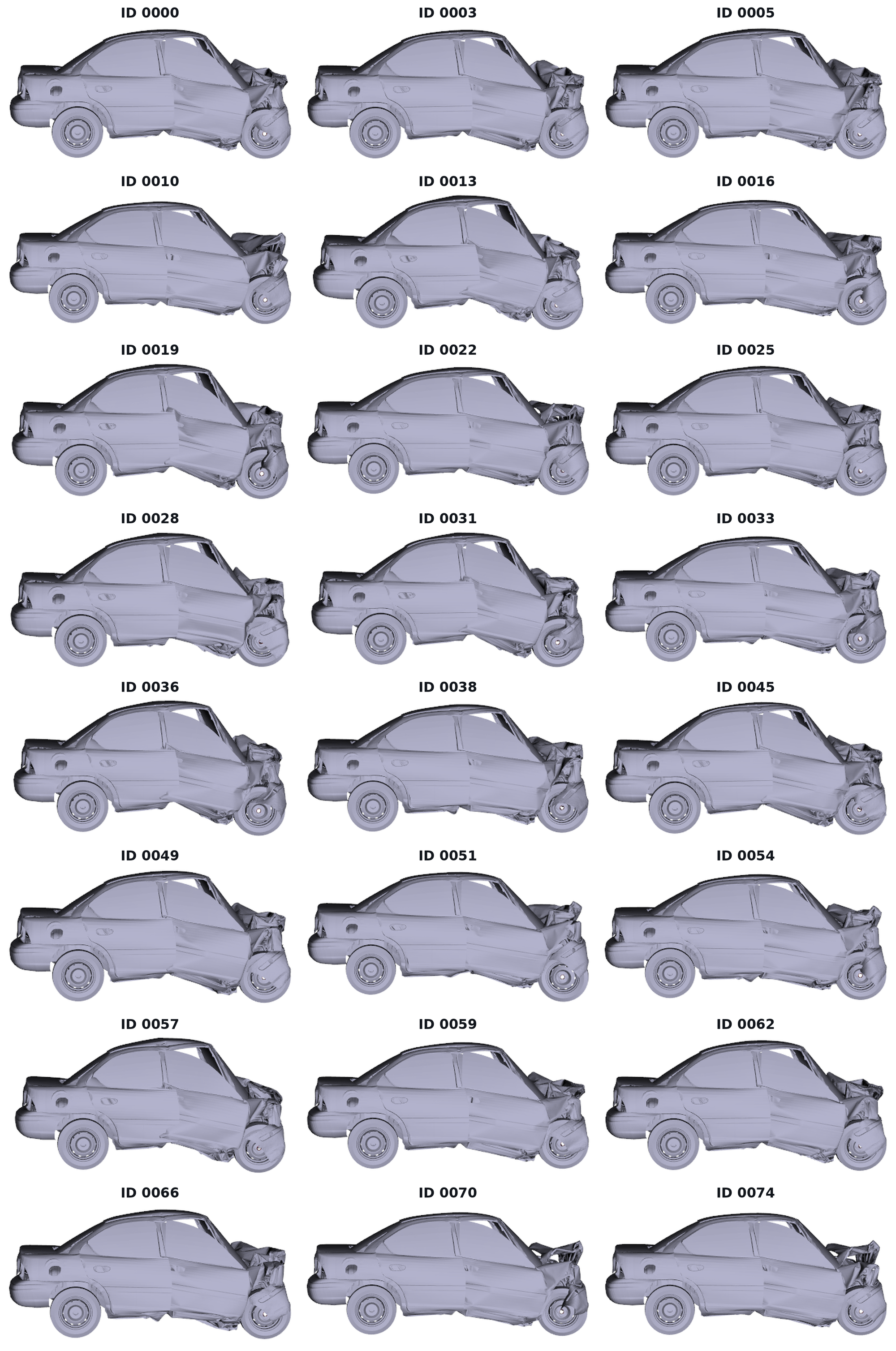}
    \caption{Representative deformed configurations from the Dodge Neon campaign shown in side view. The displayed cases emphasize variations in crush mode, wheel intrusion, and overall body deformation.}
    \label{fig:neon_campaign_side}
\end{figure}

\begin{figure}[h!]
    \centering
    \includegraphics[
        width=\linewidth,
        height=0.95\textheight,
        keepaspectratio
    ]{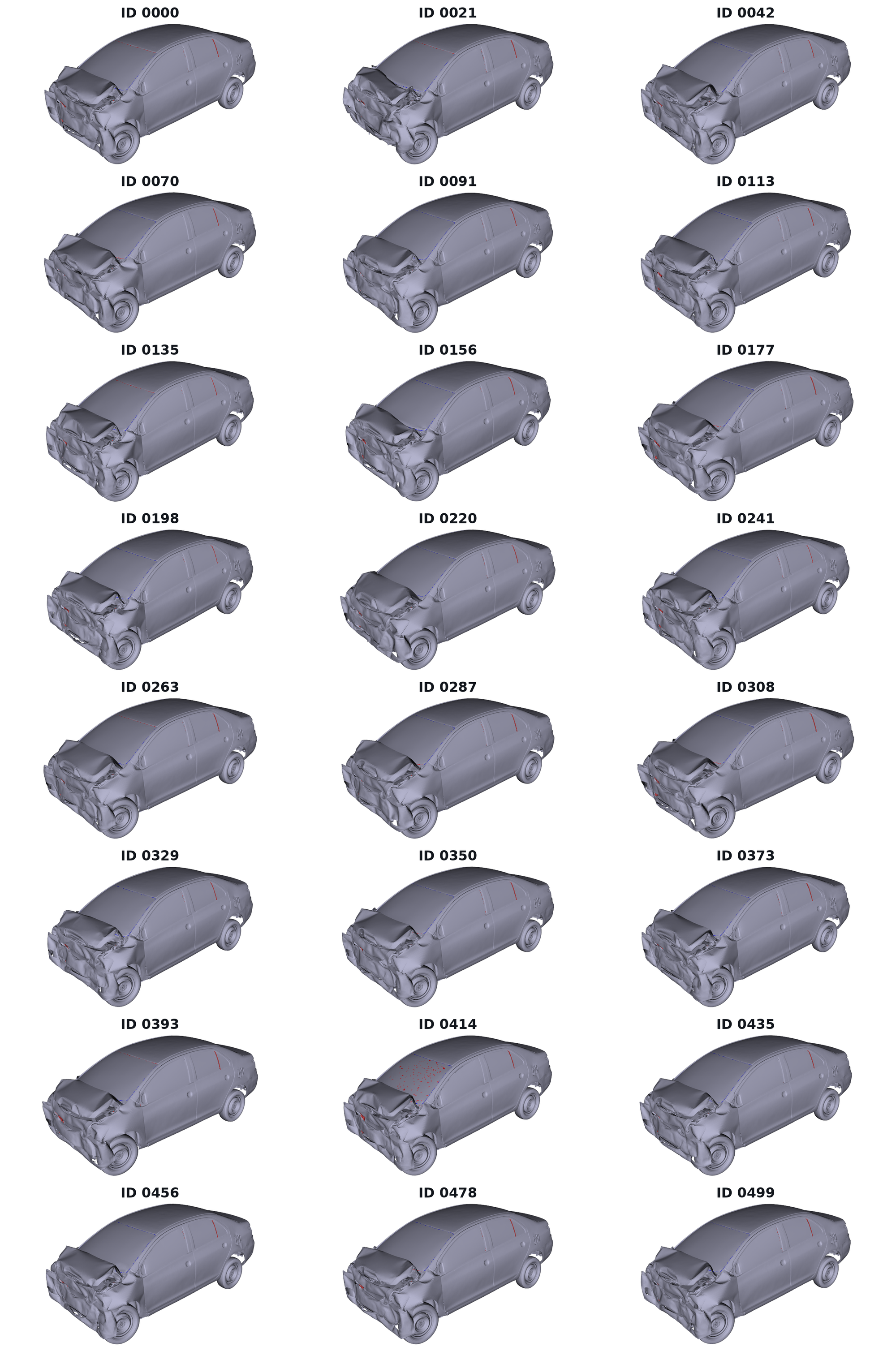}
    \caption{Representative deformed configurations from the Toyota Yaris campaign shown in isometric view. The gallery illustrates the diversity of full-vehicle crash responses across the sampled design space in \textsc{CarCrashNet}.}
    \label{fig:yaris_campaign_iso}
\end{figure}

\begin{figure}[h!]
    \centering
    \includegraphics[
        width=\linewidth,
        height=0.95\textheight,
        keepaspectratio
    ]{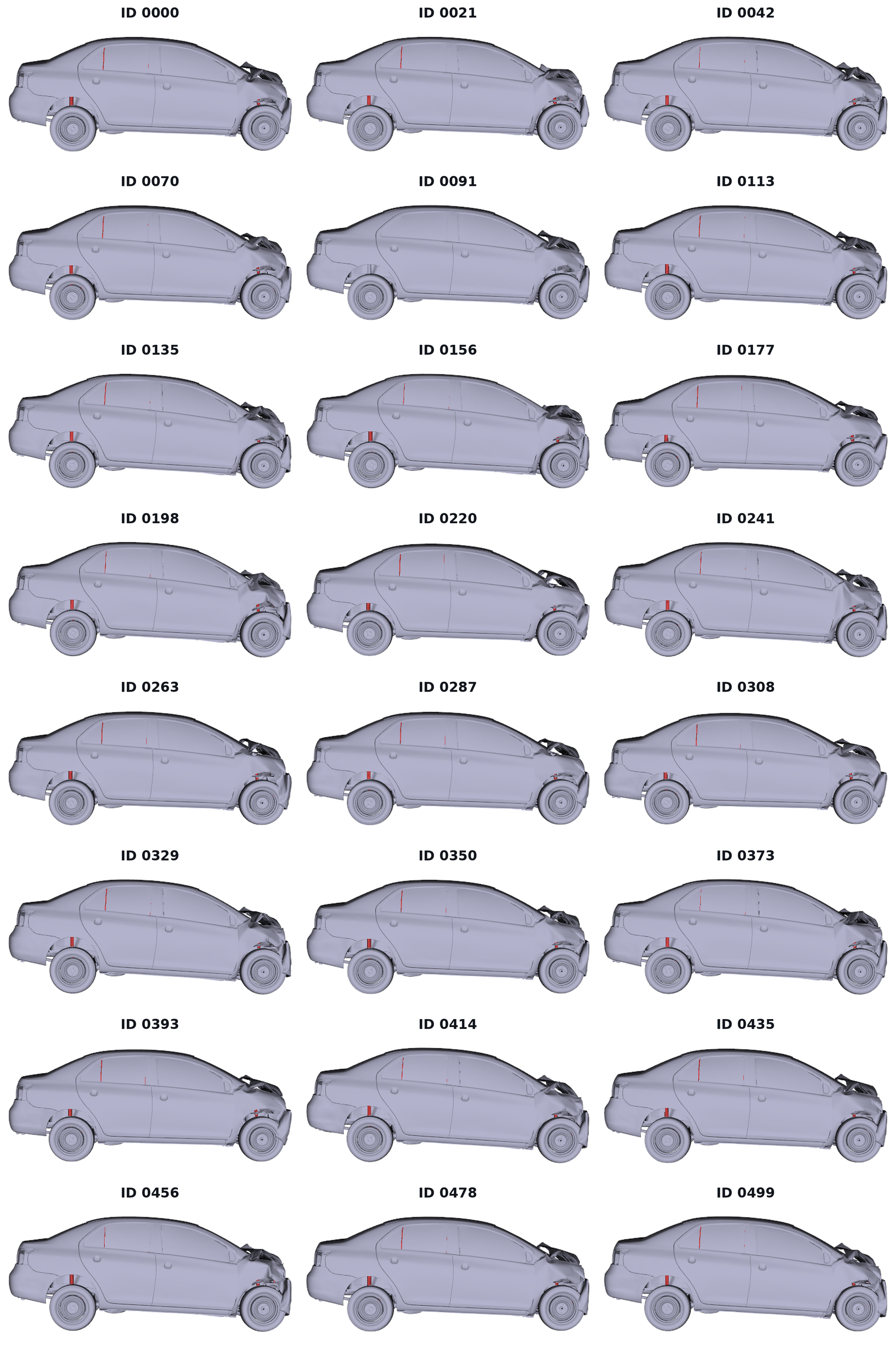}
    \caption{Representative deformed configurations from the Toyota Yaris campaign shown in side view. The selected cases highlight variations in front-end crush progression and global structural deformation across the dataset.}
    \label{fig:yaris_campaign_side}
\end{figure}

\begin{figure}[h!]
    \centering
    \includegraphics[
        width=\linewidth,
        height=0.95\textheight,
        keepaspectratio
        ]{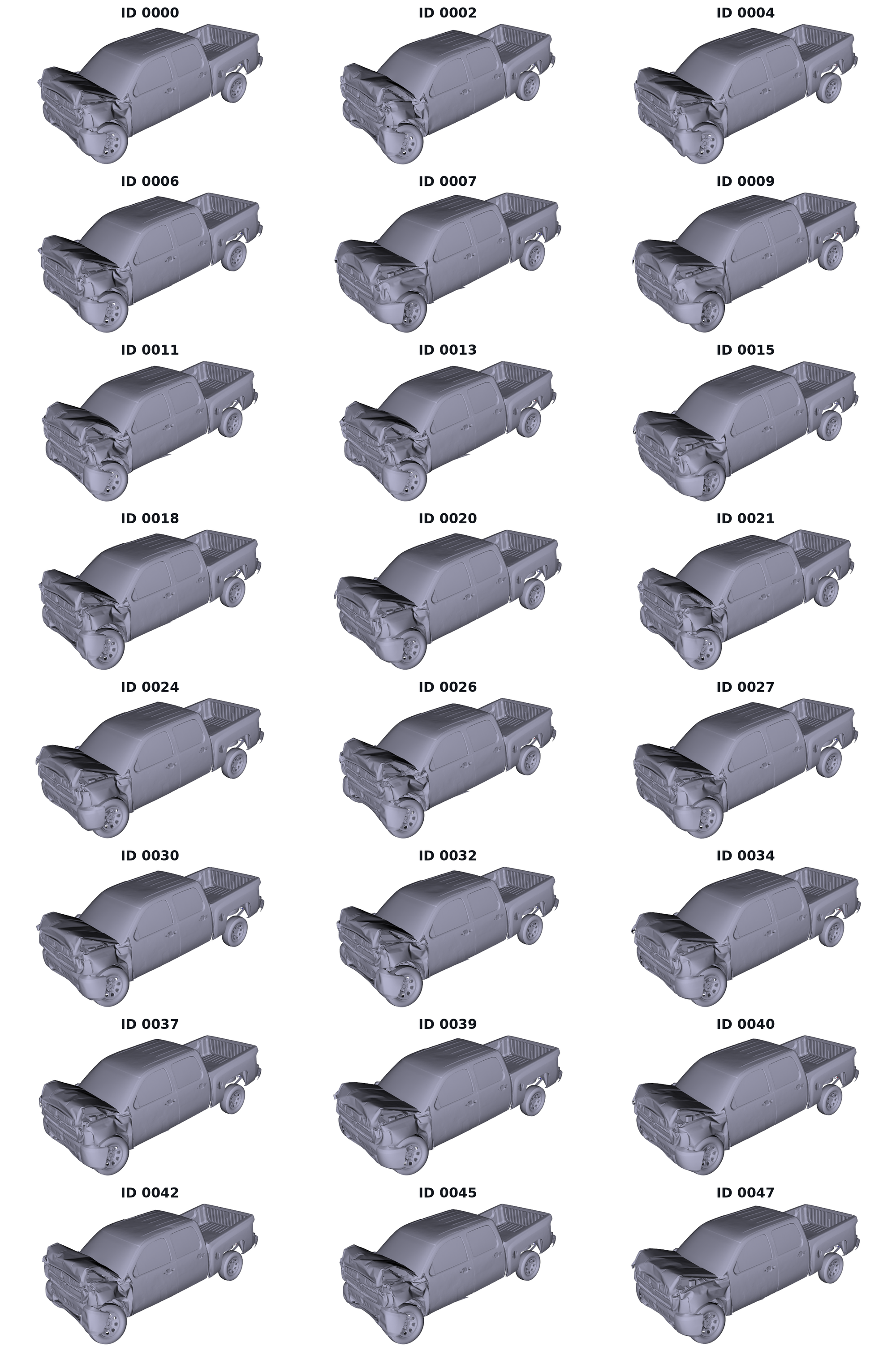}
    \caption{Representative deformed configurations from the Chevrolet Silverado campaign shown in isometric view. The gallery highlights the diversity of crash responses for the pickup-truck platform under the sampled impact conditions.}
    \label{fig:silverado_campaign_iso}
\end{figure}

\begin{figure}[h!]
    \centering
    \includegraphics[
        width=\linewidth,
        height=0.95\textheight,
        keepaspectratio
    ]{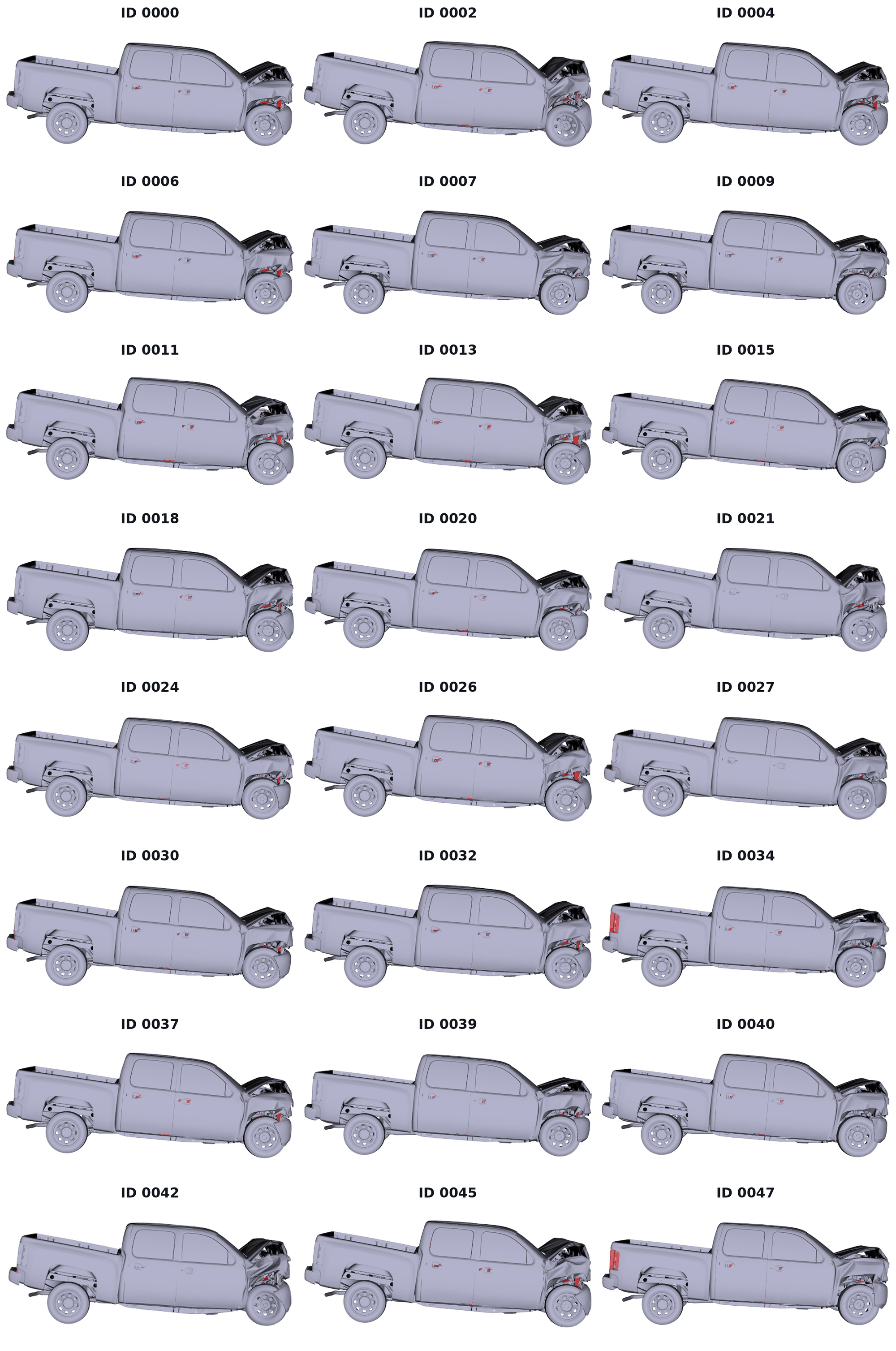}
    \caption{Representative deformed configurations from the Chevrolet Silverado campaign shown in side view. The selected cases illustrate variations in front-end structural collapse and full-vehicle deformation patterns across the dataset.}
    \label{fig:silverado_campaign_side}
\end{figure}

Figures~\ref{fig:neon_campaign_iso}--\ref{fig:silverado_campaign_side}
show representative deformed configurations from the Dodge Neon, Toyota Yaris,
and Chevrolet Silverado campaigns in both isometric and side views, illustrating
the range of vehicle-scale crash responses captured across the three datasets.

\newpage

\newpage

\newpage

\newpage

\newpage

\clearpage

\section{CrashSolver Ablations: Which Design Choices Matter?}
\label{app:crashsolver_ablation_studies}

We run a systematic architecture ablation on the Dodge Neon benchmark to
identify which CrashSolver design choices most affect performance. All runs use
the same low-resolution Dodge Neon field-prediction interface, data split,
retained-node subset, optimizer schedule, and checkpoint-selection rule. The
split contains 80 training cases, 5 validation cases, and 14 held-out test
cases. Each ablation changes one architectural parameter around the baseline
configuration, so the results should be interpreted as local sensitivity tests
rather than separately tuned leaderboard models. Metrics are computed from the
best-validation checkpoint, and lower is better for all reported errors.

\paragraph{Experiment 1: local geometry encoding.}
The first ablation studies the local component encoder, which processes retained
nodes inside each semantic structural component before global mixing. We vary
the number of slicing-attention groups and the number of local encoder layers.
As shown in Table~\ref{tab:crashsolver_neon_ablation_local_encoder}, increasing
the number of encoder slices from 16 to 32 gives the best result, reducing RMSE
from 37.135\,mm to 36.707\,mm and improving all reported metrics. In contrast,
increasing the local encoder depth to three layers degrades performance. This
suggests that better local geometric partitioning is more useful than simply
adding depth in this low-data full-vehicle setting.

\begin{table}[h!]
  \centering
  \caption{Experiment 1: local geometry encoder ablations on Dodge Neon.}
  \label{tab:crashsolver_neon_ablation_local_encoder}
  \scriptsize
  \setlength{\tabcolsep}{4pt}
  \renewcommand{\arraystretch}{1.05}
  \begin{tabular*}{\textwidth}{@{\extracolsep{\fill}} l l r r r r r @{}}
    \toprule
    \textbf{Variant} & \textbf{Change} & \textbf{RMSE} & \textbf{MAE} &
    \textbf{Rel. $L_2^x$} & \textbf{Rel. $L_2^u$} & \textbf{RMSE$_T$} \\
    \midrule
    slices32 & encoder slices 32 & \textbf{36.707} & \textbf{21.234} & \textbf{0.03080} & \textbf{0.09768} & \textbf{58.046} \\
    enc1     & encoder layers 1  & 36.892 & 21.254 & 0.03097 & 0.09832 & 59.108 \\
    base     & baseline          & 37.135 & 21.541 & 0.03118 & 0.09909 & 59.060 \\
    slices08 & encoder slices 8  & 37.165 & 21.502 & 0.03120 & 0.09911 & 59.000 \\
    enc3     & encoder layers 3  & 37.612 & 21.811 & 0.03161 & 0.10074 & 59.443 \\
    \bottomrule
  \end{tabular*}
\end{table}
\FloatBarrier

\paragraph{Experiment 2: global structural mixing.}
The second ablation studies the global component mixer, which exchanges
information across semantic components after local encoding. We vary the number
of global transformer layers and the number of global attention heads while
keeping the local encoder fixed. Table~\ref{tab:crashsolver_neon_ablation_global_mixer}
shows that the global mixer is useful but less sensitive than the local encoder:
three global layers provide a modest improvement over the two-layer baseline,
whereas changing the number of attention heads from 2 to 8 has almost no effect.
This indicates that cross-component information exchange matters, but the exact
attention-head count is not the dominant factor in this benchmark.

\begin{table}[h!]
  \centering
  \caption{Experiment 2: global mixer ablations on Dodge Neon.}
  \label{tab:crashsolver_neon_ablation_global_mixer}
  \scriptsize
  \setlength{\tabcolsep}{4pt}
  \renewcommand{\arraystretch}{1.05}
  \begin{tabular*}{\textwidth}{@{\extracolsep{\fill}} l l r r r r r @{}}
    \toprule
    \textbf{Variant} & \textbf{Change} & \textbf{RMSE} & \textbf{MAE} &
    \textbf{Rel. $L_2^x$} & \textbf{Rel. $L_2^u$} & \textbf{RMSE$_T$} \\
    \midrule
    global3 & global layers 3 & \textbf{36.919} & 21.420 & \textbf{0.03098} & \textbf{0.09829} & \textbf{59.022} \\
    global1 & global layers 1 & 36.987 & \textbf{21.388} & 0.03105 & 0.09856 & 59.290 \\
    heads2  & global heads 2  & 37.118 & 21.553 & 0.03117 & 0.09903 & 59.119 \\
    base    & baseline        & 37.135 & 21.541 & 0.03118 & 0.09909 & 59.060 \\
    heads8  & global heads 8  & 37.136 & 21.568 & 0.03118 & 0.09909 & 59.052 \\
    \bottomrule
  \end{tabular*}
\end{table}
\FloatBarrier

\paragraph{Experiment 3: model capacity.}
The third ablation tests whether larger latent or decoder dimensions improve
the model. Table~\ref{tab:crashsolver_neon_ablation_capacity} shows that
capacity is not monotonic. Reducing the latent width to 96 gives the best RMSE,
MAE, relative errors, and final-frame RMSE, while increasing the latent width to
192 or reducing the decoder width to 128 degrades performance. This suggests
that, for the available Dodge Neon training set, moderate capacity regularizes
better than simply widening the model.

\begin{table}[h!]
  \centering
  \caption{Experiment 3: capacity ablations on Dodge Neon.}
  \label{tab:crashsolver_neon_ablation_capacity}
  \scriptsize
  \setlength{\tabcolsep}{4pt}
  \renewcommand{\arraystretch}{1.05}
  \begin{tabular*}{\textwidth}{@{\extracolsep{\fill}} l l r r r r r @{}}
    \toprule
    \textbf{Variant} & \textbf{Change} & \textbf{RMSE} & \textbf{MAE} &
    \textbf{Rel. $L_2^x$} & \textbf{Rel. $L_2^u$} & \textbf{RMSE$_T$} \\
    \midrule
    latent096 & latent dim. 96      & \textbf{36.951} & \textbf{21.416} & \textbf{0.03101} & \textbf{0.09837} & \textbf{58.756} \\
    base      & baseline            & 37.135 & 21.541 & 0.03118 & 0.09909 & 59.060 \\
    dec384    & decoder hidden 384  & 37.240 & 21.458 & 0.03128 & 0.09956 & 59.165 \\
    latent192 & latent dim. 192     & 37.610 & 21.723 & 0.03161 & 0.10072 & 59.343 \\
    dec128    & decoder hidden 128  & 37.724 & 22.316 & 0.03166 & 0.10046 & 60.043 \\
    \bottomrule
  \end{tabular*}
\end{table}
\FloatBarrier

\paragraph{Experiment 4: crash conditioning and part information.}
The fourth ablation tests crash-specific auxiliary inputs: dense part-ID
embeddings, impact/positional features, and learned contact-event tokens. As
reported in Table~\ref{tab:crashsolver_neon_ablation_conditioning}, increasing
the part-embedding dimension to 32 gives the best RMSE and MAE, while an
8-dimensional part embedding gives the best final-frame RMSE and relative
displacement error. Adding more contact tokens also improves over the baseline,
whereas removing contact tokens gives the worst final-frame RMSE in this group.
These results indicate that part identity and contact-event conditioning provide
small but consistent benefits for modeling crash load transfer.

\begin{table}[h!]
  \centering
  \caption{Experiment 4: crash conditioning and part-information ablations on Dodge Neon.}
  \label{tab:crashsolver_neon_ablation_conditioning}
  \scriptsize
  \setlength{\tabcolsep}{4pt}
  \renewcommand{\arraystretch}{1.05}
  \begin{tabular*}{\textwidth}{@{\extracolsep{\fill}} l l r r r r r @{}}
    \toprule
    \textbf{Variant} & \textbf{Change} & \textbf{RMSE} & \textbf{MAE} &
    \textbf{Rel. $L_2^x$} & \textbf{Rel. $L_2^u$} & \textbf{RMSE$_T$} \\
    \midrule
    partemb32 & part embedding dim. 32 & \textbf{36.983} & \textbf{21.386} & \textbf{0.03105} & 0.09870 & 58.304 \\
    partemb8  & part embedding dim. 8  & 37.010 & 21.558 & 0.03106 & \textbf{0.09857} & \textbf{58.278} \\
    contact16 & contact tokens 16      & 37.074 & 21.420 & 0.03112 & 0.09886 & 58.536 \\
    base      & baseline               & 37.135 & 21.541 & 0.03118 & 0.09909 & 59.060 \\
    pe32      & positional dim. 32     & 37.156 & 21.559 & 0.03120 & 0.09912 & 58.657 \\
    pe08      & positional dim. 8      & 37.193 & 21.546 & 0.03123 & 0.09927 & 58.549 \\
    contact0  & contact tokens 0       & 37.233 & 21.652 & 0.03127 & 0.09946 & 59.642 \\
    \bottomrule
  \end{tabular*}
\end{table}
\FloatBarrier

\paragraph{Overall findings.}
Across the four ablation groups, the strongest improvement comes from the local
geometry encoder: increasing the number of encoder slices gives the best
overall ablation result in Table~\ref{tab:crashsolver_neon_ablation_local_encoder}.
The global mixer ablations in Table~\ref{tab:crashsolver_neon_ablation_global_mixer}
show smaller changes, suggesting that the baseline already captures most of the
needed cross-component interaction. The capacity results in
Table~\ref{tab:crashsolver_neon_ablation_capacity} show that larger models are
not automatically better, and the conditioning results in
Table~\ref{tab:crashsolver_neon_ablation_conditioning} show that part and
contact information help at the margin. Overall, these results support a
conservative CrashSolver configuration: use enough local slicing and global
mixing to represent crash load paths, include part and contact conditioning, but
avoid indiscriminately increasing model capacity without additional training
data.

\clearpage

\section{Concurrent Dataset Evaluation}
\label{app:concurrent_suv_eval}

The \textsc{CarCrashNet} vehicle-scale dataset and the SHIFT-Crash dataset\footnote{\url{https://huggingface.co/datasets/luminary-shift/SHIFT-Crash}}
were developed concurrently.  SHIFT-Crash is not part of the primary
\textsc{CarCrashNet} dataset contribution, so we keep its results out of the
main paper.  We include them here only for completeness and to provide an
additional external check of the same CrashSolver implementation on a
separately generated crash dataset.

SHIFT-Crash varies geometry and material parameters across thousands of
crash simulations rather than holding one vehicle topology fixed.  The public
dataset describes a 5{,}183-run generated design space with 376{,}593-node
source meshes, 13 output frames from 0--120\,ms, and six geometric/material
parameters.  Because the original SHIFT-Crash release does not provide an official
train/validation/test split, we define a fixed 80/10/10\% partition for our
evaluation and release the corresponding split file through Harvard Dataverse.
The converted dataset table used in our experiments contains 5{,}110 usable
runs, with 504 held-out test runs.  We report mean absolute position error, RMSE,
relative position error $L_2^x$, relative displacement error $L_2^u$, and
RMSE at 60\,ms.  Lower is better for all metrics. 

\paragraph{Dataset limitations.}
SHIFT-Crash is useful as an external check, but it is not a complete substitute
for the \textsc{CarCrashNet} dataset. Its design space is generated
from a single model family (coarse Toyota Yaris), so all samples share the same underlying
vehicle topology and frontal load-path architecture. The six reported design
variables modify global geometric offsets and front-rail properties, including
front-rail/crush-zone length, cabin length, vehicle width, roof height,
front-rail shell thickness, and front-rail yield strength. While these variables
are meaningful for parametric geometry variation, they do not provide the same
level of cross-platform structural diversity as \textsc{CarCrashNet}, which
includes three distinct public vehicle models spanning two passenger cars and a
pickup truck with different meshes, part hierarchies, material inventories, and
front-end architectures.

A second limitation is that SHIFT-Crash uses a fixed crash boundary condition.
It does not vary impact speed, contact setup, wall or pole configuration, or
other loading parameters, which restricts the range of crash severities and
contact regimes observed by the learning models. In contrast, \textsc{CarCrashNet}
explicitly varies impact velocity and crash-safety-relevant structural
thickness groups, and its component-scale bumper-beam dataset further varies
pole diameter, lateral offset, geometry, and material properties. This makes
\textsc{CarCrashNet} better suited for evaluating whether models generalize
across both structural design changes and loading-condition changes.

Finally, SHIFT-Crash primarily releases an undeformed mesh together with field
targets in a processed learning format, whereas \textsc{CarCrashNet} is designed
as a full simulation-data release. For each valid case, \textsc{CarCrashNet}
provides design metadata, solver histories, field trajectories, derived scalar
quantities, split files, and validation diagnostics, enabling both direct ML
benchmarking and independent physical auditing of the simulations. To our
knowledge, SHIFT-Crash does not report a comparable suite of ML baseline
results, so we use it here only as an external robustness check for the same
CrashSolver implementation.

\begin{table*}[h!]
  \centering
  \scriptsize
  \caption{SHIFT-Crash dataset 504-run held-out test results. Lower is better. Rows are ranked by mean RMSE. Here $L_2^x$ denotes relative position error and $L_2^u$ denotes relative displacement error.}
  \label{tab:concurrent_suv_completed}
  \setlength{\tabcolsep}{3pt}
  \renewcommand{\arraystretch}{1.05}
  \begin{tabular*}{\textwidth}{@{\extracolsep{\fill}} c l c c c c c c @{}}
    \toprule
    \textbf{Rank} &
    \textbf{Model} &
    \textbf{Runs} &
    \textbf{MAE [mm]} &
    \textbf{RMSE [mm]} &
    \textbf{Rel. $L_2^x$} &
    \textbf{Rel. $L_2^u$} &
    \textbf{RMSE@60ms [mm]} \\
    \midrule
    1 & CrashSolver PartID small & 504 & \textbf{3.736} & \textbf{6.442} & \textbf{0.00479} & \textbf{0.02251} & \textbf{5.521} \\
    2 & Transolver               & 504 & 4.016 & 6.749 & 0.00501 & 0.02357 & 5.783 \\
    3 & CrashSolver PartID DDP2  & 504 & 3.997 & 6.778 & 0.00504 & 0.02370 & 5.713 \\
    4 & GeoTransolver            & 504 & 4.663 & 7.828 & 0.00581 & 0.02746 & 6.493 \\
    5 & CrashSolver PartID+Plus  & 504 & 5.136 & 8.317 & 0.00618 & 0.02910 & 7.594 \\
    6 & CrashSolver PartID       & 504 & 5.794 & 9.258 & 0.00688 & 0.03237 & 8.721 \\
    \bottomrule
  \end{tabular*}
\end{table*}

Table~\ref{tab:concurrent_suv_completed} reports completed 504-run held-out
evaluations.  All rows use the same 80/10/10\% split and predict 12 future
displacement frames from the undeformed mesh.  The CrashSolver variant names
identify controlled implementation choices: \textit{PartID} uses learned dense
embeddings of the original finite-element \texttt{PART\_ID} values;
\textit{PartID+Plus} adds an enhanced local point-feature backbone before
hierarchical mixing; \textit{DDP2} denotes a longer distributed-data-parallel
training run of the PartID model; and \textit{small} denotes a lower-capacity
PartID model with fewer latent channels, slices, and attention heads.  The
CrashSolver PartID small model is the current full-test leader on this split,
improving over Transolver from 4.016 to 3.736\,mm MAE, from 6.749 to
6.442\,mm RMSE, from 5.783 to 5.521\,mm RMSE@60\,ms, and from 0.00501 to
0.00479 relative position error $L_2^x$.  Its displacement-normalized error
$L_2^u$ is also slightly lower than Transolver, improving from 0.02357 to
0.02251.

\clearpage

\section{Uncertainty and Statistical Significance}
\label{app:statistical_significance}

We quantify uncertainty in the field-prediction benchmarks using paired
statistics over the fixed held-out test cases. For each dataset and model, we
compute per-case RMSE and MAE from the saved evaluation outputs. We then
estimate 95\% confidence intervals for each model mean using nonparametric
bootstrap resampling with 10{,}000 replicates. For pairwise model comparisons,
we preserve case-level pairing and bootstrap the per-case RMSE differences.
We also report paired win rates, two-sided sign-flip permutation tests, and
two-sided Wilcoxon signed-rank tests. These intervals quantify uncertainty over
the fixed held-out test sets, not variability across training seeds.

\begin{table*}[h!]
  \centering
  \caption{
    Bootstrap 95\% confidence intervals for mean RMSE and MAE on the
    field-prediction benchmarks. Lower is better. Intervals are computed over
    held-out test cases and measure fixed-test-set uncertainty.
  }
  \label{tab:statistical_significance_mean_ci}
  \scriptsize
  \setlength{\tabcolsep}{3pt}
  \renewcommand{\arraystretch}{1.05}
  \begin{tabular*}{\textwidth}{@{\extracolsep{\fill}} l l r r r @{}}
    \toprule
    \textbf{Dataset} & \textbf{Model} & \textbf{Runs} &
    \textbf{RMSE mean [95\% CI]} & \textbf{MAE mean [95\% CI]} \\
    \midrule
    \multirow{4}{*}{Toyota Yaris}
      & CrashSolver   & 50 & \textbf{21.769} [19.056, 24.562] & 13.507 [11.559, 15.555] \\
      & GeoTransolver & 50 & 21.773 [19.038, 24.550] & \textbf{13.359} [11.375, 15.394] \\
      & FIGConvUNet   & 50 & 21.910 [19.240, 24.615] & 13.576 [11.617, 15.582] \\
      & Transolver    & 50 & 22.583 [19.952, 25.300] & 14.049 [12.119, 16.057] \\
    \midrule
    \multirow{4}{*}{Dodge Neon}
      & CrashSolver   & 25 & \textbf{32.763} [26.477, 39.031] & \textbf{18.036} [14.646, 21.335] \\
      & Transolver    & 25 & 33.947 [27.916, 39.841] & 18.678 [15.291, 22.082] \\
      & FIGConvUNet   & 25 & 34.044 [28.111, 40.155] & 18.850 [15.551, 22.201] \\
      & GeoTransolver & 25 & 34.403 [27.981, 41.112] & 18.973 [15.410, 22.576] \\
    \midrule
    \multirow{4}{*}{Chevrolet Silverado}
      & CrashSolver   & 15 & \textbf{61.536} [51.462, 72.584] & \textbf{37.753} [29.733, 45.875] \\
      & GeoTransolver & 15 & 79.230 [70.135, 88.988] & 45.366 [37.263, 53.989] \\
      & Transolver    & 15 & 83.971 [75.038, 93.999] & 47.510 [39.427, 55.914] \\
      & FIGConvUNet   & 15 & 102.747 [94.838, 111.417] & 62.405 [55.922, 69.422] \\
    \midrule
    \multirow{6}{*}{SHIFT-Crash}
      & CrashSolver PartID small & 504 & \textbf{6.442} [6.271, 6.616] & \textbf{3.736} [3.639, 3.832] \\
      & Transolver               & 504 & 6.749 [6.564, 6.944] & 4.016 [3.905, 4.132] \\
      & CrashSolver PartID DDP2  & 504 & 6.778 [6.599, 6.961] & 3.997 [3.900, 4.098] \\
      & GeoTransolver            & 504 & 7.828 [7.612, 8.045] & 4.663 [4.537, 4.795] \\
      & CrashSolver PartID+Plus  & 504 & 8.317 [8.157, 8.478] & 5.136 [5.038, 5.234] \\
      & CrashSolver PartID       & 504 & 9.258 [9.066, 9.456] & 5.794 [5.670, 5.924] \\
    \bottomrule
  \end{tabular*}
\end{table*}

Table~\ref{tab:statistical_significance_mean_ci} shows that CrashSolver obtains
the lowest mean RMSE on all four benchmarks. On Toyota Yaris, the margin over
GeoTransolver is negligible, indicating that the two methods are effectively
tied on mean RMSE. On Dodge Neon, Chevrolet Silverado, and SHIFT-Crash, the
CrashSolver advantage is larger, especially for the Silverado dataset, where
the vehicle topology and structural load paths are more complex.

\begin{table*}[h!]
  \centering
  \caption{
    Paired RMSE significance tests against the leading CrashSolver row for each
    dataset. The difference is reported as CrashSolver minus the comparison
    model, so negative values favor CrashSolver. Win rate is the fraction of
    held-out cases for which CrashSolver has lower RMSE.
  }
  \label{tab:statistical_significance_paired_tests}
  \scriptsize
  \setlength{\tabcolsep}{3pt}
  \renewcommand{\arraystretch}{1.05}
  \begin{tabular*}{\textwidth}{@{\extracolsep{\fill}} l l r r r r r @{}}
    \toprule
    \textbf{Dataset} & \textbf{Comparison} & \textbf{Pairs} &
    \textbf{Mean diff. [95\% CI]} & \textbf{Win rate} &
    \textbf{Perm. $p$} & \textbf{Wilcoxon $p$} \\
    \midrule
    Toyota Yaris & CrashSolver vs. GeoTransolver & 50 & -0.004 [-0.178, 0.176] & 0.59 & 0.9628 & 0.7158 \\
    Toyota Yaris & CrashSolver vs. FIGConvUNet   & 50 & -0.142 [-0.227, -0.059] & 0.66 & 0.0027 & 0.0039 \\
    Toyota Yaris & CrashSolver vs. Transolver    & 50 & -0.814 [-0.963, -0.652] & 0.84 & 0.0001 & $<10^{-4}$ \\
    \midrule
    Dodge Neon & CrashSolver vs. Transolver    & 25 & -1.185 [-1.552, -0.895] & 1.00 & 0.0001 & $<10^{-4}$ \\
    Dodge Neon & CrashSolver vs. FIGConvUNet   & 25 & -1.282 [-1.850, -0.723] & 0.68 & 0.0006 & 0.0010 \\
    Dodge Neon & CrashSolver vs. GeoTransolver & 25 & -1.641 [-3.180, -0.033] & 0.60 & 0.0564 & 0.0028 \\
    \midrule
    Chevrolet Silverado & CrashSolver vs. GeoTransolver & 15 & -17.694 [-20.949, -13.652] & 1.00 & 0.0001 & 0.0001 \\
    Chevrolet Silverado & CrashSolver vs. Transolver    & 15 & -22.435 [-25.967, -17.972] & 0.93 & 0.0003 & 0.0001 \\
    Chevrolet Silverado & CrashSolver vs. FIGConvUNet   & 15 & -41.211 [-45.853, -35.690] & 1.00 & 0.0001 & 0.0001 \\
    \midrule
    SHIFT-Crash & PartID small vs. Transolver              & 504 & -0.307 [-0.554, -0.070] & 0.53 & 0.0144 & 0.0271 \\
    SHIFT-Crash & PartID small vs. PartID DDP2             & 504 & -0.336 [-0.446, -0.228] & 0.62 & 0.0001 & $<10^{-4}$ \\
    SHIFT-Crash & PartID small vs. GeoTransolver           & 504 & -1.386 [-1.571, -1.206] & 0.76 & 0.0001 & $<10^{-4}$ \\
    SHIFT-Crash & PartID small vs. PartID+Plus             & 504 & -1.875 [-1.991, -1.759] & 0.93 & 0.0001 & $<10^{-4}$ \\
    SHIFT-Crash & PartID small vs. PartID                  & 504 & -2.816 [-2.962, -2.673] & 0.97 & 0.0001 & $<10^{-4}$ \\
    \bottomrule
  \end{tabular*}
\end{table*}

The paired tests in Table~\ref{tab:statistical_significance_paired_tests}
confirm the same pattern. On Toyota Yaris, CrashSolver and GeoTransolver are
statistically tied: the paired confidence interval contains zero and both
paired tests are non-significant. CrashSolver is nevertheless significantly
better than FIGConvUNet and Transolver on the same test set. On Dodge Neon,
CrashSolver is significantly better than Transolver and FIGConvUNet, while the
GeoTransolver comparison is mixed: the bootstrap interval and Wilcoxon test
favor CrashSolver, but the sign-flip permutation test is slightly above
$p=0.05$. On Chevrolet Silverado and SHIFT-Crash, the leading CrashSolver
variant is significantly better than every listed comparison across the paired
bootstrap intervals, permutation tests, and Wilcoxon tests.

\clearpage

\section{Bumper-Beam Pole-Impact Crashworthiness Dataset}
\label{sec:bumper-dataset}

\subsection{Structural Model}
\label{sec:dataset:model}

\begin{figure}[h!]
    \centering
    \includegraphics[width=\linewidth]{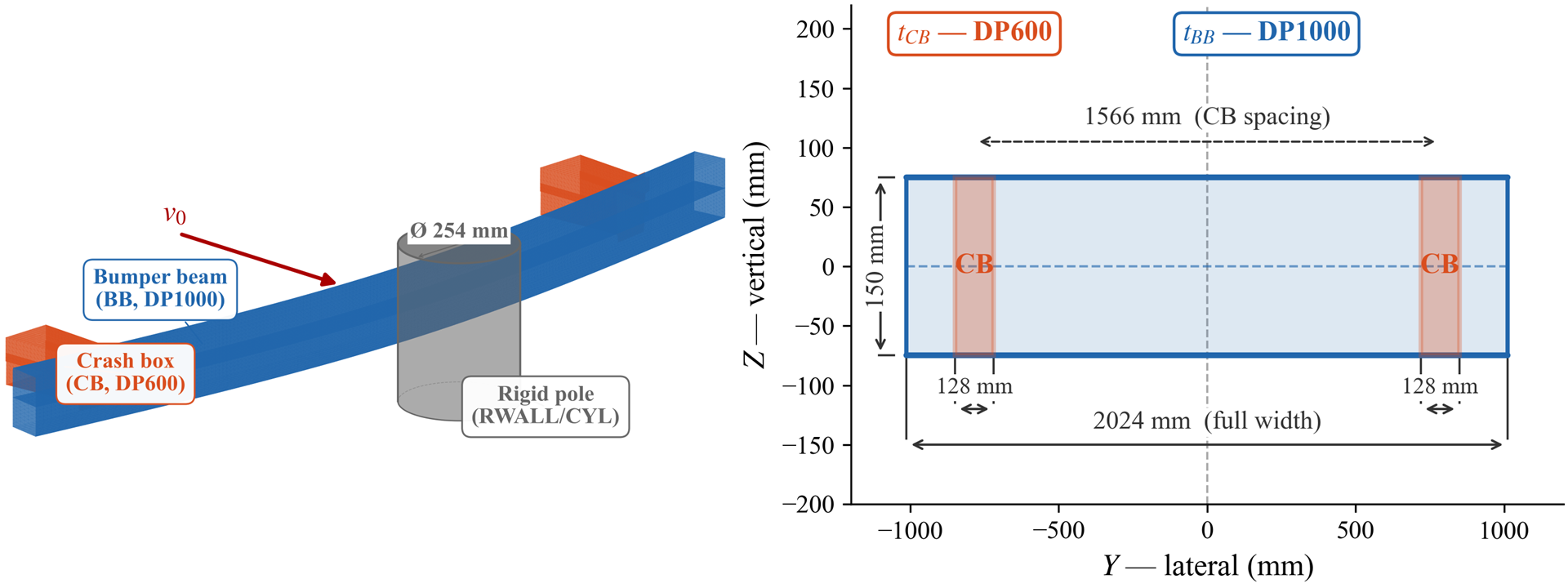}
    \caption{Geometry and loading setup of the simplified bumper beam assembly used for dataset generation. The left panel shows the isometric view of the pole-impact configuration used in the finite element simulations, where the bumper beam impacts a rigid cylindrical pole. The right panel shows the front-view schematic of the bumper beam and crash-box assembly, highlighting the main geometric dimensions and material regions, with the bumper beam modeled using \texttt{DP1000} and the crash boxes modeled using \texttt{DP600}.}
    \label{fig:bumper_beam_setup}
\end{figure}

The finite-element model used for dataset generation represents a bumper beam
assembly of a simplified frontal crash structure, consisting of two primary
structural members (see Figure~\ref{fig:bumper_beam_setup}): the
\emph{crash-box} (energy-absorbing deformable tube, DP600 grade steel) and
the \emph{bumper beam} (cross-car bending member, DP1000 grade steel),
together forming a 14-part assembly with bilateral symmetry.
The mesh comprises 13,832 nodes and 13,598 quadrilateral shell elements
discretised across 20 parts, modelled with full numerical integration (Ishell~=~4)
using the Belytschko--Tsay shell formulation~\citep{belytschko1984explicit}.
The model is solved with the explicit finite-element code
OpenRadioss using 8 MPI ranks per simulation.
The simulation terminates at $t = 100$\,ms, outputting 101 animation frames
at $\Delta t_{\text{anim}} = 1$\,ms intervals.

\subsection{Material Models}
\label{sec:dataset:materials}

Both steel grades are modelled with the
\emph{Johnson--Cook} elasto-viscoplastic constitutive law~\citep{johnson1983cook}, which describes the equivalent flow stress as:
\begin{equation}
  \bar{\sigma} = \bigl(A + B\,\bar{\varepsilon}_p^{\,n}\bigr)
    \bigl(1 + C\ln\dot{\bar{\varepsilon}}^*\bigr)
    \bigl(1 - T^{*m}\bigr),
  \label{eq:jc}
\end{equation}
where $\bar{\varepsilon}_p$ is the accumulated plastic strain, and
$\dot{\bar{\varepsilon}}^* = \dot{\bar{\varepsilon}}_p / \dot{\varepsilon}_0$
is the dimensionless strain rate.
In the generated material cards, rate and thermal factors are disabled, so the
evaluated response uses the quasi-static, isothermal hardening law
$\bar{\sigma} = A + B\,\bar{\varepsilon}_p^n$ rather than relying on a
degenerate setting of the full multiplicative form.
Both materials share the same elastic constants:
$E = 210$\,GPa, $\nu = 0.30$,
$\rho = 7.85 \times 10^{-6}$\,kg/mm$^3$.

The Johnson--Cook yield-strength parameter $A$ is the initial yield stress; in
dataset tables and machine-learning feature lists we denote the same physical
quantity by $\sigma_y$. The yield-strength parameter is varied as a design
parameter across the dataset. The remaining material parameters are summarized
in Table~\ref{tab:materials}. The baseline hardening coefficients $B = A$ and
$n = A/\text{GPa}$ are scaled proportionally with yield so that the tangent
modulus scales consistently with typical DP steel experimental data. For
Iterations~1--2 of the design-of-experiments
(Section~\ref{sec:dataset:execution}), no element deletion is activated
(\texttt{EPS\_p\_max}~$=0$), so those simulations capture the full
large-deformation post-buckling response without mesh erosion. Iteration~3
introduces a small failure-strain regularisation
$\bar\varepsilon^p_{\text{fail}}=0.05$ to prevent abnormal terminations on
severely distorted elements at the extremes of the design space; the resulting
eroded element population stays below 0.5\,\% of the structural mesh and does
not change the macroscopic load--deflection response.

\begin{table}[t]
  \centering
  \caption{
    Material parameters held constant across all simulations.
    The yield parameter $A$ is varied per the design of experiments
    (Table~\ref{tab:doe_params}).
  }
  \label{tab:materials}
  \small
  \begin{tabular}{lcccccc}
    \toprule
    \textbf{Grade} & \textbf{Part} & $E$ (GPa) & $\nu$ & $\rho$ (kg/mm$^3$) & $B$ & $n$ \\
    \midrule
    DP600  & Crash-box   & 210 & 0.30 & $7.85\!\times\!10^{-6}$ & $=A$ & $=A/\text{GPa}$ \\
    DP1000 & Bumper beam & 210 & 0.30 & $7.85\!\times\!10^{-6}$ & $=A$ & $=A/\text{GPa}$ \\
    \bottomrule
  \end{tabular}
\end{table}

\subsection{Boundary Conditions}
\label{sec:dataset:bc}

\paragraph{Initial velocity.}
The entire bumper beam assembly is given a uniform translational initial
velocity $V_x = -v$ in the global $-X$ direction,
applied to a node group comprising all structural parts (crash-box, bumper
beam, spotwelds, and associated brackets).
The rigid-wall pole is stationary; the structure impacts the pole.

\paragraph{Rigid wall.}
The impactor is a cylindrical rigid wall with its axis
aligned with the global $Z$-direction, centred at $(x_c, y_p, 0)$.
The pole diameter $D_p$ is a design parameter; the $X$-position of the pole
centre is automatically adjusted to guarantee a minimum clearance of
$\Delta_{\text{gap}} = 10$\,mm between the pole surface and the bumper front
face at time $t = 0$:
\begin{equation}
  x_c = -\!\left(\left|x_{\text{bumper}}\right| + \tfrac{D_p}{2} + \Delta_{\text{gap}}\right),
  \label{eq:pole_x}
\end{equation}
where $x_{\text{bumper}} \approx -40$\,mm is the $X$-coordinate of the
bumper front surface in the reference configuration.
The lateral offset $y_p$ is also a design parameter, enabling centred and
offset impacts to be represented within a single model.
Coulomb friction between the structure and the pole is set to $\mu = 0.20$.

\paragraph{Contact.}
Self-contact within the assembly is handled by a penalty-based single-surface
contact, covering all structural parts, with automatic gap calculation (\texttt{Igap=2})
and consistent tangent stiffness (\texttt{Istf=4}).
Spotweld connections between the crash-box and bumper beam flanges are modelled
with tied-contact constraints.

\paragraph{Constraints.}
Rigid-wall rear nodes (representing the vehicle body behind the assembly)
are restrained in translation ($Y$, $Z$) and all three rotations, simulating attachment of the assembly to a rigid-body
representation of the vehicle front structure.

\paragraph{Timestep control.}
A nodal critical time-step scheme is used with
scale factor $\alpha = 0.90$ and a minimum time-step floor of
$\Delta t_{\min} = 10^{-3}$\,ms.
Nodal mass scaling is allowed above the floor, with added mass routinely
below 0.03\,\% of total structural mass across the dataset.

\subsection{Design of Experiments}
\label{sec:dataset:doe}

\paragraph{Parameter space.}
Seven design parameters are varied across simulations, as summarized in
Table~\ref{tab:doe_params}. These parameters were selected to span an
engineering-relevant design space for crashworthiness analysis. Two gauge
parameters, crash-box thickness $t_{\mathrm{cb}}$ and bumper-beam thickness
$t_{\mathrm{bb}}$, and two material yield-strength parameters independently
control the structural capacity of the energy-absorbing crash boxes and the
cross-car bumper beam. Three loading and contact parameters, impact velocity
$v$, pole diameter $D_p$, and lateral pole offset $y_p$, vary the impact
severity and contact configuration.

\begin{table}[h!]
  \centering
  \caption{
    Design parameters and their ranges. All parameters are sampled using a
    scrambled Sobol sequence subject to the engineering feasibility constraints
    in Eq.~\eqref{eq:constraints}.
  }
  \label{tab:doe_params}
  \small
  \setlength{\tabcolsep}{4pt}
  \begin{tabular}{llccc}
    \toprule
    \textbf{Parameter} & \textbf{Description} & \textbf{Min} & \textbf{Max} & \textbf{Grid step} \\
    \midrule
    $v$              & Impact velocity        & 2.0 mm/ms (7.2 km/h)  & 15.0 mm/ms (54 km/h) & 0.5 mm/ms \\
    $t_{\mathrm{cb}}$  & Crash-box thickness    & 1.0 mm                & 3.0 mm               & 0.1 mm    \\
    $t_{\mathrm{bb}}$  & Bumper-beam thickness  & 1.0 mm                & 3.0 mm               & 0.1 mm    \\
    $A_{\mathrm{DP600}}$  & Crash-box yield parameter & 0.150 GPa        & 0.600 GPa            & 25 MPa    \\
    $A_{\mathrm{DP1000}}$ & Bumper-beam yield parameter & 0.250 GPa     & 1.000 GPa            & 25 MPa    \\
    $D_p$            & Pole diameter          & 100 mm                & 500 mm               & 10 mm     \\
    $y_p$            & Lateral pole offset    & 0 mm                  & 800 mm               & 25 mm     \\
    \bottomrule
  \end{tabular}
\end{table}

\paragraph{Quasi-random sampling.}
Candidate design points are drawn from a scrambled Sobol sequence
(dimension~$=7$)~\citep{owen1998scrambling} using
\texttt{scipy.stats.qmc.Sobol}. Sobol sampling provides low-discrepancy
coverage of the design space and improves coverage of parameter-space corners
for a fixed simulation budget. Raw samples in $[0,1]^7$ are linearly mapped to
the physical ranges and rounded to the engineering grid steps in
Table~\ref{tab:doe_params}.

\paragraph{Engineering feasibility constraints.}
Not all combinations of gauge and yield strength produce physically meaningful
or manufacturable structural configurations. Very thin sheets with low yield
strength can lead to unstable local deformation near connections, while very
thick sheets with very high yield strength fall outside the intended design
window for automotive dual-phase steels. We therefore enforce
\begin{equation}
\begin{gathered}
\begin{aligned}
  0.8 &\leq t_{\mathrm{bb}}\sqrt{A_{\mathrm{DP1000}}} \leq 2.5, \\
  0.6 &\leq t_{\mathrm{cb}}\sqrt{A_{\mathrm{DP600}}}  \leq 2.0,
\end{aligned}
\\
A_{\mathrm{DP1000}} \geq A_{\mathrm{DP600}}.
\end{gathered}
\label{eq:constraints}
\end{equation}
where the gauge-strength products are in
mm$\cdot\!\sqrt{\mathrm{GPa}}$. The first two constraints restrict the
structural capacity of the bumper beam and crash boxes to a practical design
window, while the third enforces that the bumper beam, which acts as the primary
bending member, is made from a stronger steel grade than the crash boxes.

\paragraph{Geometric pre-screening.}
Before solver execution, candidate pole placements are checked for initial
geometric intersection with the bumper mesh. For each candidate, we compute the
minimum Euclidean distance in the $XY$-plane between the pole center and the
$N_{\mathrm{str}}$ structural node positions in the reference configuration:
\begin{equation}
  \delta_{\min} =
  \min_{j=1,\ldots,N_{\mathrm{str}}}
  \sqrt{(x_j - x_c)^2 + (y_j - y_p)^2}.
\end{equation}
A candidate is accepted only if $\delta_{\min} \geq D_p/2$, ensuring that the
pole surface does not intersect the structure at $t=0$. Combined with the
pole-placement rule in Eq.~\eqref{eq:pole_x}, this pre-screening step removes
geometrically invalid designs before running the explicit solver.

\paragraph{Industry-inspired test points.}
In addition to the Sobol-sampled designs, we include mandatory anchor points
representing common crashworthiness regimes and design-space extremes, including
low-speed bumper impact, frontal impact speeds near 50--54 km/h, lightest and
heaviest gauge settings, smallest and largest pole diameters, and maximum
lateral offset. These anchor cases are injected ahead of the Sobol sequence.

\subsection{Campaign Execution}
\label{sec:dataset:execution}

The corpus was generated in three iterations, each motivated by an
information-gap audit of the previous batch. The final released corpus contains
14{,}742 finite-element simulations with seven design inputs and scalar
crashworthiness targets, together with the corresponding metadata needed for
phase-aware ablation studies.

\paragraph{Iteration 1: Baseline.}
The first iteration established the initial Sobol-sampled design space using
seed~42. Candidate designs were launched in batches of 24 concurrent jobs with
8 MPI ranks per simulation on a 192-core workstation. This iteration defined
the baseline simulation workflow, material setup, pole-placement strategy, and
post-processing pipeline used for the subsequent dataset-generation campaigns.

\paragraph{Iteration 2: Diversity expansion.}
The second iteration expanded the design coverage using a new Sobol seed
(seed~2025) and activated the per-sample geometric pre-filter described in
Section~\ref{sec:dataset:doe}. This removed invalid initial pole--bumper
intersections before solver execution and enabled reliable sampling over the
full pole-diameter range $D_p\in[100,500]$\,mm. Iterations~1 and~2 use the
original Johnson--Cook material card with \texttt{EPS\_p\_max}~$=0$,
corresponding to no element deletion, so they capture the full large-deformation
post-buckling response.

\paragraph{Iteration 3: High-resolution sweep.}
The third iteration densified the Sobol pre-grid for impact velocity and pole
diameter, increasing the number of unique values for $v$ and $D_p$ relative to
the earlier iterations. This sweep also introduced a small failure-strain
regularisation,
\texttt{EPS\_p\_max}~$=\bar{\epsilon}^{p}_{\text{fail}}=0.05$, to prevent
abnormal terminations for severely distorted elements at the extremes of the
design space. The high-resolution sweep retains the same seven-feature
predictive interface as Iterations~1--2.

\subsection{Physics Validation}
\label{sec:dataset:validation}

Release rows are selected through an automated quality screen applied to solver
logs and global energy histories
($\Delta T_{\text{out}}=0.1$\,ms). The screen removes abnormal terminations,
invalid initial contact configurations, unstable energy histories, excessive
hourglass response, and severe time-step collapse; mass scaling is retained as
a diagnostic field rather than silently clipped. Some low-energy or rebound
cases finish before the nominal 100\,ms output horizon, so final simulated time
is reported in Table~\ref{tab:dataset_stats} rather than imposed as a fixed
eligibility value. After this audit, the released bumper-beam corpus contains
\textbf{14{,}742} finite-element simulations; discarded launch attempts and
failed quality-control rows are not used in the benchmark splits.

\begin{table}[h]
  \centering
  \caption{
    Input and output statistics for the 14{,}742 finite-element release rows.
    Energies and forces use the OpenRadioss
    $\text{(t, mm, ms)}$ unit system, which yields kJ for energy and
    N for force.
    $E_{\text{kin},0}$: initial kinetic energy;
    $E_{\text{int}}^{\max}$: peak internal energy during crash;
    $W_p^{\max}$: peak cumulative plastic work;
    $F_p^{\max}$: peak pole reaction/contact force (occupant-loading
    proxy and bumper-level analogue of $F_{\mathrm{wall}}^{\max}$);
    $a^{\max}$: peak rigid-body deceleration of the impacting assembly;
    $\eta_{\text{KE}} = 1-E_{\text{kin}}(T_f)/E_{\text{kin}}(0)$.
  }
  \label{tab:dataset_stats}
  \small
  \setlength{\tabcolsep}{6pt}
  \begin{tabular}{lrrrr}
    \toprule
    \textbf{Quantity} & \textbf{Min} & \textbf{Median} & \textbf{Max} & \textbf{Unit} \\
    \midrule
    \multicolumn{5}{l}{\emph{DoE inputs}}\\
    Impact velocity $v$         &   2.0  &   8.0  &   15.0 & mm/ms \\
                                 &   7.2  &  28.8  &   54.0 & km/h  \\
    Crash-box thickness $t_{\text{cb}}$ &  1.00  &  2.05  &  3.00 & mm    \\
    Bumper-beam thickness $t_{\text{bb}}$ &  1.00 &  1.95  &  3.00 & mm   \\
    DP600 yield $\sigma_{y,\mathrm{DP600}}$ &  0.150&  0.350 &  0.600 & GPa  \\
    DP1000 yield $\sigma_{y,\mathrm{DP1000}}$ & 0.250& 0.675 &  1.000 & GPa  \\
    Pole diameter $D_p$         &  100   &  250   &  500   & mm    \\
    Pole lateral offset $y_p$   &    0   &  400   &  800   & mm    \\
    \midrule
    \multicolumn{5}{l}{\emph{Response targets}}\\
    $E_{\text{kin},0}$          &  1.02  &  16.43 &  59.69 & kJ    \\
    $E_{\text{int}}^{\max}$     &  0.66  &  12.49 &  60.07 & kJ    \\
    $W_p^{\max}$                &  0.20  &  11.16 & 941.5\textsuperscript{$\dagger$} & kJ \\
    $F_p^{\max}$                &  44    &  935   &  9{,}802 & N   \\
    $a^{\max}$                  &  189   & 12{,}408 & 82{,}076 & mm/ms$^2$ \\
    $\eta_{\text{KE}}$          &  5.6   &  97.4  & 100.0 & \%    \\
    \midrule
    Energy balance error $|E_{\text{err}}|$ &  0.0 &  1.4 &  5.0 & \% \\
    Hourglass ratio              &  0.0  &  0.0  &  0.0  & \%    \\
    Added mass                   &  0.02 &  0.02 & 13.49\textsuperscript{$\ddagger$} & \% \\
    Final simulation time        & 86.3  & 99.9  & 100.0 & ms    \\
    \bottomrule
  \end{tabular}\\
  {\footnotesize $^\dagger$ The reported maximum is dominated by a single retained edge-case simulation with severe element distortion and mass scaling; for physically well-resolved cases, cumulative plastic work tracks peak internal energy, as reflected by the comparable medians (11.16~kJ vs 12.49~kJ).\;
  $\ddagger$~The median added mass is near the diagnostic floor; one retained
  phase-3 row is a documented mass-scaling outlier.}
\end{table}

\paragraph{Dataset statistics.}
Table~\ref{tab:dataset_stats} summarises the design inputs and the
post-processed scalar response targets over the released corpus.
The two-order-of-magnitude spread in peak internal energy
(0.66--60.07\,kJ) and the wide range of KE absorption
(5.6--100\,\%) confirm that the DoE samples both light
''elastic bounce'' and severe plastic collapse regimes, providing the
diversity required for surrogate-model generalisation.

\paragraph{Physical interpretation.}
The median KE absorption of 97.4\,\% confirms that the majority of
simulations represent complete structural arrest within the
100\,ms window, consistent with bumper-level impacts at urban speeds.
The median plastic-work fraction $W_p^{\max}/E_{\text{int}}^{\max}=0.89$
agrees with DP-steel experimental data, which typically exhibits
80--92\,\% plastic dissipation in progressive crush
events~\citep{tarigopula2006}.
The 33.5\,\% of simulations with $\eta_{\text{KE}}<95\,\%$ correspond
primarily to low-velocity ($v<3.5$\,mm/ms) or large-pole/large-offset
configurations in which the structure rebounds elastically before fully
arresting --- a physically correct behaviour that is important for the
surrogate to reproduce.
Hourglass and mass-scaling diagnostics are essentially flat across the
corpus (medians at the floor of measurement), apart from the documented
mass-scaling outlier reported in Table~\ref{tab:dataset_stats}.  The low
median values reflect the use of the
Flanagan--Belytschko viscous--stiffness control with $q_h=0.10$ and the
$\Delta t_{\min}=10^{-3}$\,ms time-step floor. 

\paragraph{Released dataset.}
The final bumper-beam corpus contains $\mathbf{14{,}742}$ finite-element
simulations distributed through the public release table
\texttt{bumper\_beam\_master.csv}. Each record contains the seven design inputs
listed in Table~\ref{tab:doe_params}, the post-processed scalar response metrics
reported in Table~\ref{tab:dataset_stats}, the pole-placement geometry, and a
stable identifier of the form \texttt{sim\_XXXXX}. This identifier links each
row to its corresponding VTKHDF mesh-state archive and JSON sidecar containing
the design parameters and boundary conditions used for that simulation. The
iteration index, \texttt{phase}~$\in\{1,2,3\}$, is preserved as metadata for
ablation studies but is not used as a model input.

\begin{figure}[p]
    \centering
    \includegraphics[
        width=\linewidth,
        height=0.86\textheight,
        keepaspectratio
    ]{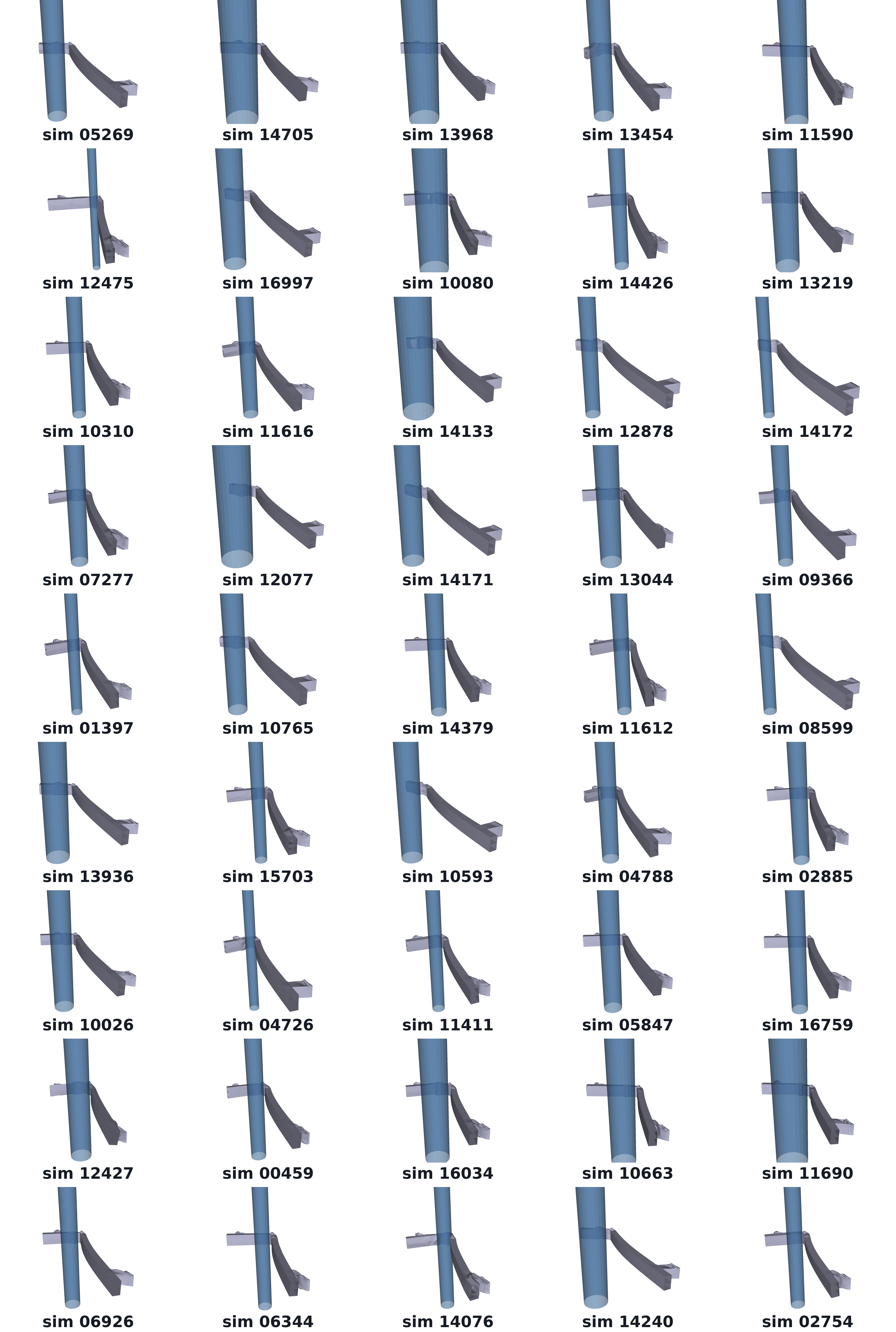}
    \caption{Representative bumper-beam pole-impact configurations sampled from the design space. Each panel shows a different simulation instance with variations in pole diameter, lateral offset, and structural geometry, illustrating the diversity of contact conditions and deformation modes covered by our dataset.}
    \label{fig:gallery_iso}
\end{figure}

\begin{figure}[p]
    \centering
    \includegraphics[
        width=\linewidth,
        height=0.86\textheight,
        keepaspectratio
    ]{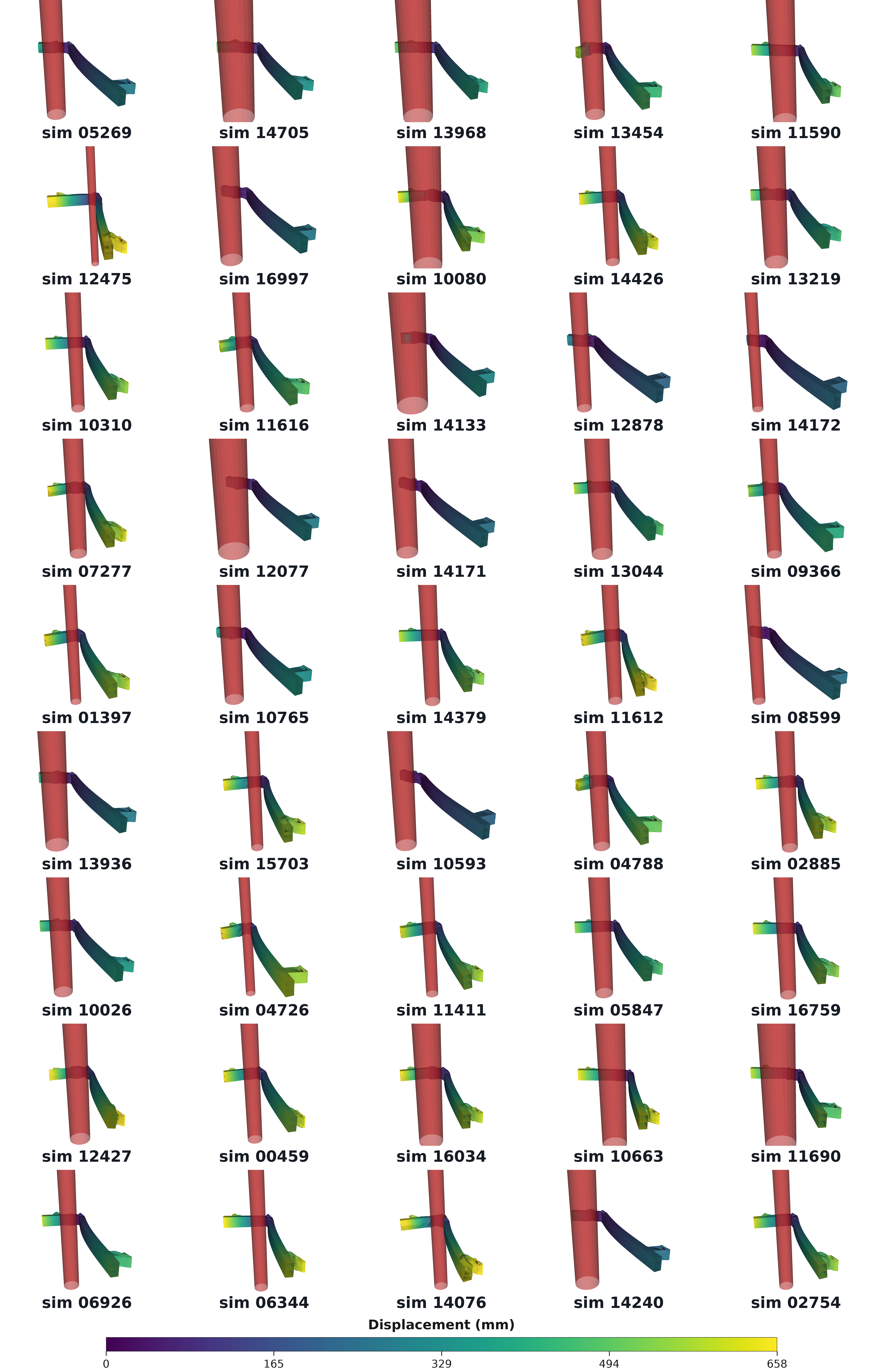}
    \caption{Representative deformed configurations from the bumper-beam dataset used in \textsc{CarCrashNet}. The figure shows a diverse set of pole-impact simulations sampled from the design space, with displacement magnitude visualized on the deformed structures. The examples highlight variations in global bending, local crushing near the contact region, and the resulting range of structural responses captured across the dataset.}
    \label{fig:bumper_beam_gallery}
\end{figure}

Figures~\ref{fig:gallery_iso} and~\ref{fig:bumper_beam_gallery} illustrate the diversity of bumper-beam pole-impact simulations used for dataset generation, showing variations in contact configuration, deformation mode, and displacement response across the sampled design space.

\clearpage

\clearpage

\clearpage

\section{Bumper-Beam Machine-Learning Benchmark}
\label{sec:ml-benchmark}

Engineering data is often most naturally represented in tabular form, where a compact set of design variables is paired with a corresponding set of quantities of interest. Such parametric representations are particularly valuable in engineering design, as they enable systematic exploration of design spaces, facilitate the identification of correlations between inputs and performance metrics, and provide interpretable summaries of complex simulation campaigns. In the present bumper-beam study, each simulation can be described by a number of design parameters together with a set of scalar crashworthiness responses, yielding a structured tabular dataset that is well suited for surrogate modeling. We therefore train and compare twelve regression models on the validated bumper-beam dataset to predict five crashworthiness metrics from these design parameters. The goal is to obtain a fast, differentiable surrogate that can approximate explicit finite-element simulations within an outer design and optimization loop.

\subsection{Dataset and Splits}
\label{subsec:ml-dataset}

The training corpus consists of 14,742 fully-validated simulations. Each record is described by seven structural/loading features
and five scalar response targets (Table~\ref{tab:ml-features-targets}).
For the bumper-beam task, $F_p^{\max}$ denotes the peak pole reaction or
contact force and is the bumper-level analogue of the vehicle-scale
$F_{\mathrm{wall}}^{\max}$.
The dataset is split once, with seed~42, into a 70/15/15\%
train/validation/test partition with \num{2212} held-out test simulations.
Validation is used only for early stopping and AutoML model selection; all
reported metrics come from the held-out test set. The split indices are
persisted to
\texttt{bumper\_beam\_ml\_splits.json} so that every model sees
identical data.

\begin{table}[h!]
\centering
\small
\caption{Input features (7) and prediction targets (5).}
\label{tab:ml-features-targets}
\begin{tabular}{lll}
\toprule
\multicolumn{3}{l}{\textbf{DoE features}} \\
\midrule
Impact velocity          & $v$                          & \si{\milli\metre\per\milli\second} \\
Crash-box thickness      & $t_{\mathrm{cb}}$           & \si{\milli\metre} \\
Bumper-beam thickness    & $t_{\mathrm{bb}}$           & \si{\milli\metre} \\
DP600 yield stress       & $\sigma_{y,\mathrm{DP600}}$ & \si{\giga\pascal} \\
DP1000 yield stress      & $\sigma_{y,\mathrm{DP1000}}$& \si{\giga\pascal} \\
Pole diameter            & $D_p$                       & \si{\milli\metre} \\
Pole lateral offset      & $y_p$                       & \si{\milli\metre} \\
\midrule
\multicolumn{3}{l}{\textbf{Response targets}} \\
\midrule
Peak pole contact force  & $F_p^{\max}$                & \si{\newton} \\
Peak deceleration        & $a^{\max}$                  & \si{\milli\metre\per\milli\second\squared} \\
Peak internal energy     & $E_{\mathrm{int}}^{\max}$   & \si{\kilo\joule} \\
Peak plastic work        & $W_p^{\max}$                & \si{\kilo\joule} \\
Kinetic-energy absorbed  & $\eta_{\mathrm{KE}}$        & \si{\percent} \\
\bottomrule
\end{tabular}
\end{table}

\subsection{Model Suite}
\label{subsec:ml-models}

We evaluate four model families, chosen to span the full complexity
spectrum from a closed-form linear fit to an in-context foundation model.
Hyperparameters are held fixed across targets to emulate a practitioner's
zero-tuning workflow; differences therefore reflect inductive bias,
not tuning effort.

\paragraph{Linear and kernel baselines.}
Ridge and Lasso regression with $\alpha=1$ and $\alpha=10^{-2}$
respectively provide a reference for the portion of variance that is
explained by a global linear response. A $k$-nearest-neighbors
($k=10$, inverse-distance weights) regressor and a radial-basis-function
support-vector regressor ($C=10$) probe locally smooth structure.

\paragraph{Neural network.}
A two-layer fully-connected MLP with hidden widths
$(128, 128)$, ReLU activations, and Adam optimisation is trained for up
to \num{400} epochs with early stopping on the validation set.

\paragraph{Tree ensembles.}
Random Forest and Extra Trees (each with \num{400} estimators) capture
non-linear feature interactions without explicit tuning.

\paragraph{Gradient boosting.}
XGBoost, LightGBM, and CatBoost are trained with up to \num{2000} rounds,
learning rate $\eta = 0.03$, maximum depth~6 (or \num{63} leaves for
LightGBM), and early stopping after~50 rounds of no improvement on the
validation set.

\paragraph{Foundation model.}
TabPFN~v2~\citep{hollmann2022tabpfn}, a transformer pre-trained in-context
on synthetic tabular distributions, is evaluated in its regression mode
on a single NVIDIA A100 GPU with
\texttt{ignore\_pretraining\_limits=True} to handle the
\num{10319}-sample training set.

\paragraph{AutoML.}
AutoGluon Tabular~\citep{erickson2020autogluon} is run with the
\texttt{best\_quality} preset and a 5-minute time budget per target.
The validation split is supplied as tuning data with
\texttt{use\_bag\_holdout=True}.

\subsection{Evaluation Protocol}
\label{subsec:ml-eval}

For each (model, target) pair we report the coefficient of determination
$R^2$, mean absolute error (MAE), root-mean-square error (RMSE), and
mean absolute percentage error (MAPE). Early stopping uses
only the validation split; AutoML's internal model selection uses
only train$\cup$validation. Final performance is the held-out test-set
score, which no model sees during any fitting step.

\subsection{Results}
\label{subsec:ml-results}

Table~\ref{tab:ml-r2} lists the test-set $R^2$ for every
model--target pair, with the winner in bold. The gradient-boosting trio
clearly dominates: CatBoost has the highest mean test $R^2$ at
\num{0.82}, followed by LightGBM (\num{0.82}) and XGBoost (\num{0.79}),
each winning at least one target.

\begin{table}[h!]
\centering
\small
\caption{Test-set $R^2$ for 12 models $\times$ 5 targets. Best per column in \textbf{bold};
mean over targets in the last column.}
\label{tab:ml-r2}
\begin{tabular}{lcccccc}
\toprule
Model & $F_p^{\max}$ & $\eta_{\mathrm{KE}}$ & $a^{\max}$ & $E_{\mathrm{int}}^{\max}$ & $W_p^{\max}$ & Mean \\
\midrule
Ridge         & 0.280 & 0.116 & 0.520 & 0.754 & 0.750 & 0.484 \\
Lasso         & 0.280 & 0.116 & 0.520 & 0.754 & 0.750 & 0.484 \\
KNN           & 0.609 & 0.198 & 0.686 & 0.765 & 0.747 & 0.601 \\
SVR           & 0.404 & 0.077 & 0.275 & 0.262 & 0.281 & 0.260 \\
MLP           & 0.726 & 0.281 & 0.735 & 0.795 & 0.782 & 0.664 \\
\midrule
Random Forest & 0.744 & 0.514 & 0.787 & 0.902 & 0.862 & 0.762 \\
Extra Trees   & 0.756 & 0.495 & 0.791 & 0.897 & 0.886 & 0.765 \\
\midrule
XGBoost       & 0.820 & 0.627 & 0.837 & \textbf{0.918} & 0.744 & 0.789 \\
LightGBM      & 0.817 & 0.633 & 0.835 & 0.917 & \textbf{0.889} & 0.818 \\
CatBoost      & \textbf{0.822} & \textbf{0.648} & \textbf{0.847} & 0.916 & 0.878 & \textbf{0.822} \\
\midrule
TabPFN~v2     & 0.775 & 0.333 & 0.802 & 0.884 & 0.868 & 0.732 \\
AutoGluon     & 0.745 & 0.513 & 0.790 & 0.902 & 0.851 & 0.760 \\
\bottomrule
\end{tabular}
\end{table}

\paragraph{Per-target behaviour.}
$E_{\mathrm{int}}^{\max}$ is the easiest response
(best $R^2 = 0.918$) because it is strongly tied to the initial
kinetic-energy scale. $W_p^{\max}$ is also well captured
(best $R^2 = 0.889$), although XGBoost is a clear outlier on this
target. Peak deceleration and peak pole force are harder but still
well predicted by CatBoost, with best $R^2 = 0.847$ and $0.822$,
respectively. The limiting target is $\eta_{\mathrm{KE}}$, where the best
model reaches only $R^2 = 0.648$; this response is partly saturated
and therefore concentrates useful variance in a smaller subset of the DoE
space.

\paragraph{Linear baselines.}
Ridge and Lasso recover approximately $75\%$ of the variance
for $E_{\mathrm{int}}^{\max}$ and $W_p^{\max}$, but only $12\%$ for
$\eta_{\mathrm{KE}}$ and $28\%$ for $F_p^{\max}$. This confirms that
energy-like quantities are dominated by smooth global scaling, while
contact-force and absorption-fraction metrics require nonlinear
interactions among velocity, pole position, and section stiffness.

\begin{wrapfigure}{r}{0.48\columnwidth}
    \centering
    \includegraphics[width=0.47\columnwidth]{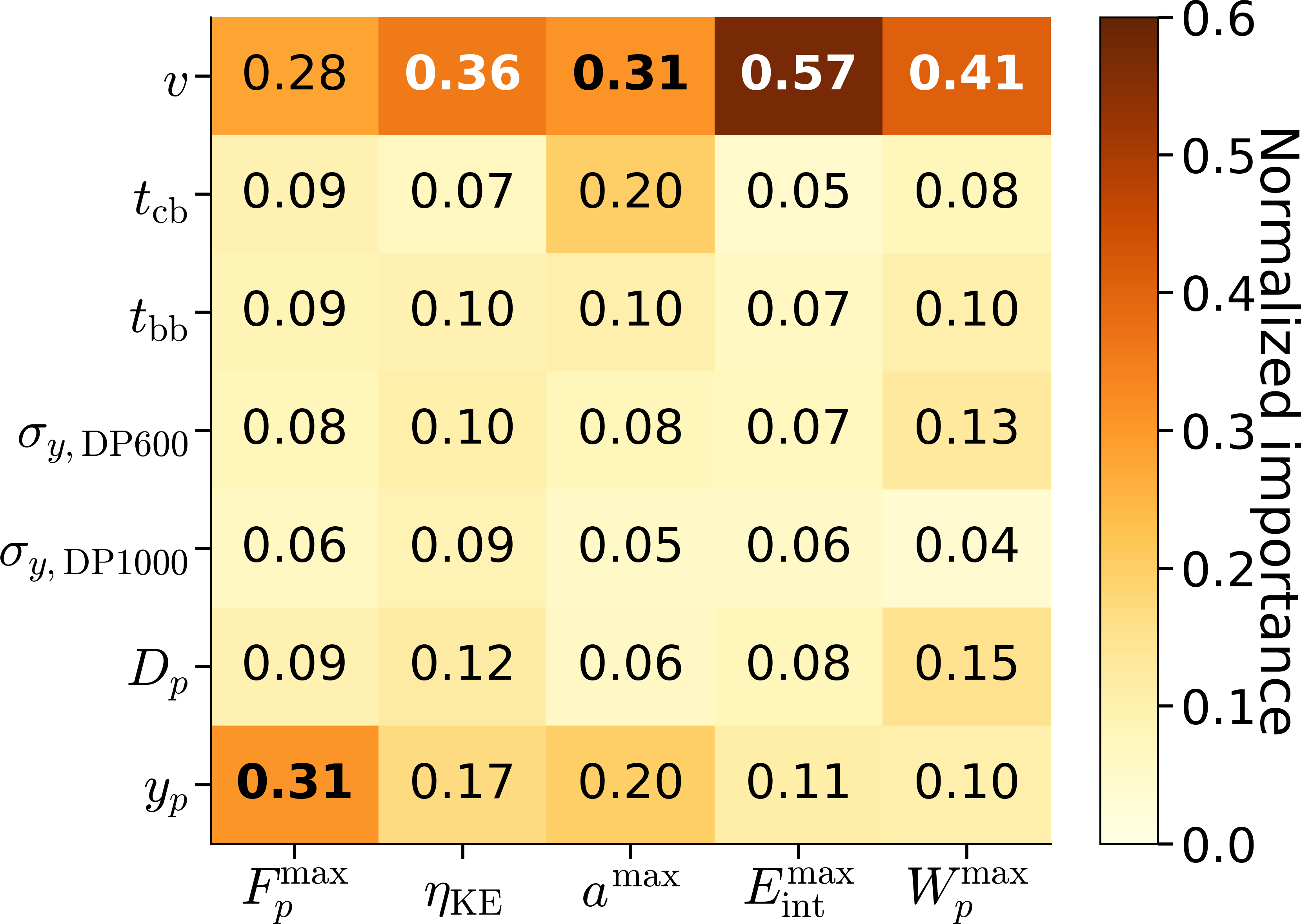}
    \caption{Column-normalized feature importance averaged over XGBoost, LightGBM, and CatBoost. The plotted values are model-internal importance scores averaged over the three boosters, then normalized independently within each target so that every target column sums to one.}
    \label{fig:ml-feature-importance}
        \vspace{-1.8em}

\end{wrapfigure}

\paragraph{Feature sensitivity.}
The normalized booster-importance map in
Fig.~\ref{fig:ml-feature-importance} matches the expected crash
physics. Impact velocity is the dominant driver for
$E_{\mathrm{int}}^{\max}$, $W_p^{\max}$, and $\eta_{\mathrm{KE}}$, where
energy scaling is the leading effect. Peak pole force is instead most
sensitive to lateral pole offset, followed by velocity, because the
offset controls how directly the pole loads the bumper beam and crash-box
load path. Peak deceleration depends on a broader mix: velocity, pole
offset, and crash-box thickness all carry comparable normalized
importance.

\paragraph{AutoML versus hand-picked boosters.}
AutoGluon's \texttt{best\_quality}
preset reaches a mean $R^2$ of \num{0.760}, close to the bagged-tree
baselines but below all three gradient boosters. The best direct booster
exceeds AutoGluon on every target by
$\Delta R^2 \in [\num{0.016}, \num{0.135}]$. On this compact
seven-feature problem, AutoML therefore acts as a useful sanity check but
does not replace a directly trained LightGBM or CatBoost model unless the
time budget is increased substantially.

\paragraph{Foundation model.}
TabPFN~v2 achieves competitive accuracy on four of five targets
($R^2 \in [\num{0.775}, \num{0.884}]$) without target-specific tuning,
which is notable because the dataset has more than \num{10000} training
points and continuous engineering responses. Its weak result on
$\eta_\text{KE}$ ($R^2=\num{0.333}$) is the main failure mode, consistent
with a generic tabular prior struggling on a clipped, boundary-saturated
response.

\subsection{Summary}
\label{subsec:ml-summary}

Gradient-boosted trees provide the most useful surrogate family for this
pole-impact benchmark. CatBoost gives the best mean accuracy
($R^2 = 0.822$), while LightGBM gives nearly the same accuracy
($R^2 = 0.818$) with roughly an order-of-magnitude lower training cost. Linear
models are adequate only for the smooth energy responses; they miss the
nonlinear contact and saturation effects that determine $F_p^{\max}$ and
$\eta_{\mathrm{KE}}$. The practical surrogate choice is therefore CatBoost when
maximum accuracy is required and LightGBM when rapid retraining inside a design
loop is more important.

This component-scale benchmark complements the vehicle-scale
\textsc{CarCrashNet} datasets rather than replacing them. The bumper-beam
setting provides a controlled design space in which geometry, material
strength, thickness, impact velocity, pole diameter, and lateral offset can be
varied systematically over thousands of simulations. This makes it well suited
for studying design-variable sensitivity, scalar crashworthiness prediction,
and fast tabular surrogate modeling. In contrast, the full-vehicle datasets
capture realistic vehicle topology, heterogeneous part interactions, and
high-resolution deformation fields, but are more expensive and harder to sample
densely. Together, the two levels provide a useful hierarchy: the bumper-beam
dataset enables broad and interpretable exploration of crash-design variables,
while the full-vehicle datasets test whether learned models can scale to
realistic structural complexity and full-field crash prediction.

\end{document}